# Fuzzy Label: From Concept to Its Application in Label Learning

Chenxi Luo[a], Zhuangzhuang Zhao[a], Zhaohong Deng[a*], Te Zhang[b]

[a] Jiangnan University, Wuxi, China
[b] Xiongan Institute of Artificial Intelligence, Hebei, China
[*] Corresponding authors: dengzhaohong@jiangnan.edu.cn

*Abstract*—**Label learning is a fundamental task in machine learning that aims to construct intelligent models using labeled data, encompassing traditional single-label and multi-label classification models. Traditional methods typically rely on logical labels, such as binary indicators (e.g., "yes/no") that specify whether an instance belongs to a given category. However, in practical applications, label annotations often involve significant uncertainty due to factors such as data noise, inherent ambiguity in the observed entities, and the subjectivity of human annotators. Therefore, representing labels using simplistic binary logic can obscure valuable information and limit the expressiveness of label learning models. To overcome this limitation, this paper introduces the concept of *fuzzy labels*, grounded in fuzzy set theory, to better capture and represent label uncertainty. We further propose an efficient fuzzy labeling method that mines and generates fuzzy labels from the original data, thereby enriching the label space with more informative and nuanced representations. Based on this foundation, we present fuzzy-label-enhanced algorithms for both single-label and multi-label learning, using the classical K-Nearest Neighbors (KNN) and multi-label KNN algorithms as illustrative examples. Experimental results indicate that fuzzy labels can more effectively characterize the real-world labeling information and significantly enhance the performance of label learning models.**

## I. INTRODUCTION

Label learning is a fundamental task in machine learning, and can be broadly categorized into two main types: 1) Single-Label learning and 2) Multi-Label Learning. In Single-Label Learning [1], [2], each input instance is associated with exactly one class label. This includes binary classification problem (e.g., 'yes/no') as well as multi-class classification problem tasks (e.g., 'Low/Medium/High'). In contrast, Multi-Label Learning [3], [4] allows each instance to be lined to multiple labels simultaneously, represented as a label vector. It is particularly useful in scenarios where the presence of multiple, co-occurring categories is common, such as image annotation.

Most existing label learning methods mainly focus on modeling the mapping from the feature space to the label space, often overlooking the latent information inherent in the labels themselves. In practical annotation, category information is often simplified into binary logical labels [5]-[7], which fails to account for the varying degrees of contribution and semantic importance of different features, restricting the representational capacity of traditional label learning. This simplification causes models to favor learning salient feature patterns that are strongly correlated with labels during training, while struggling to capture more subtle, fine-grained features that may be semantically important. As a result, the prediction model's ability can be significantly impaired.

To address the limitations of traditional logical labeling, recent research has explored soft label learning techniques [8]-[10]. These approaches transform conventional binary hard labels into soft labels represented in probabilistic form, thereby providing a more nuanced indication of an instance's degree of association with various categories. For example, Thierry Denoeux et al. [11] proposed the Possibility Labels method, which assumes that each instance has only single "true" label, and uses probability values to express the degree of compatibility between the instance and all possible labels. Building on this idea, Geng et al. [12] introduced Label Distribution Learning (LDL), which models the distribution over labels for each instance. LDL not only quantifies the description degree of each label but also captures inter-label correlations. In this framework, each instance is linked to all candidate labels, but with varying degrees of importance.

Although soft label methods like LDL alleviate some of limitations of traditional logical labeling, they remain constrained by the so-called "completeness assumption." This assumption requires that all label values for an instance must add up to one. This assumption often fails to reflect the true semantic relationships among labels in real-world scenarios. For example, as illustrated in Fig. 1(a), both "Scenery" and "Sky" may be highly descriptive labels for this given image. However, due to the normalization constraint imposed by label distribution models, increase the importance of one label inevitably reduces that of the other, which introduces a false mutual exclusivity. This limitation prevents the model from effectively capturing label co-occurrence patterns, ultimately reducing the quality and semantic richness of the label representation. Overcoming the completeness assumption remains a major challenge, as it is essential for developing more flexible and expressive label representations that better align with the meaning of real-world data.

In practice, category boundaries are often ambiguous, and the degree to which an instance belongs to a category often exhibits inherent fuzziness. For example, expressions such as "this animal looks more like a sheep" or "this assessment is basically passing" reflect partial or uncertain membership in a category. Therefore, more flexible techniques to model label uncertainty are needed. Fuzzy set theory [13], [14] offers a powerful framework for handling uncertainty and imprecision, making it a promising foundation for advancing label learning theories and methods that better align with real-world semantics.

Motivated by these insights, this paper conducts an in-depth investigation into fuzzy-label-based learning. *First,* a novel concept called Fuzzy Label grounded in fuzzy set theory is proposed. This concept leverages fuzzy sets to represent the uncertainty inherent in class boundaries, thereby enhancing the quality and expressiveness of label representations. Specifically, a membership degree – a real value within the range $[0,1]$ is used to quantify the extent to which an instance belongs to a particular category. This value is defined is defined as the fuzzy label of the instance with respect to that category. By relaxing the completeness assumption, fuzzy labels allow the use of intuitive fuzzy semantics and membership assignments, which help better model uncertainty and vagueness in category boundaries, while also capturing hidden semantic information in a more accurate and meaningful way. Secondly, since directly obtaining fuzzy labels in practice is often challenging, people have become accustomed to using simple logical labels to represent label information. Therefore, there is a pressing need for effective methods that can generate fuzzy labels for observed instances using available data. In this regard, this paper further proposes a practical fuzzy label generation method, designed to infer and extract latent uncertainty information from raw input features and corresponding logical label. The model generates fuzzy labels that better reflect the inherent fuzziness of real-world data, and enrich the information content available for learning. Thirdly, to validate the effectiveness of fuzzy labels in label learning, this paper adopts the traditional K-Nearest Neighbors (KNN) classification algorithm [15] and the Multi-Label K-Nearest Neighbors (ML-KNN) classification algorithm [16] as illustrative baselines. Based on these, we propose fuzzy label-enhanced single-label KNN learning and multi-label KNN learning methods. The proposed methods integrate fuzzy labels into the traditional KNN and ML-KNN frameworks, utilizing richer, uncertainty-aware label information during the learning process so as to enhance learning capacity and generalization performance. In sum, the main contribution of this paper are as follows:

(1) Introduction of the fuzzy label concept: To address the inherent uncertainty in real-world labeling scenarios, this paper introduces the concept of fuzzy labels, laying the foundation for a new more expressive and flexible label learning paradigm.

(2) Fuzzy label generation method: To overcome the challenge of acquiring fuzzy labels in practical settings, a feasible fuzzy label generation method is proposed for generating fuzzy labels by mining latent uncertainty from raw input data and logical labels.

(3) Fuzzy label-enhanced learning methods: To demonstrate the applicability of fuzzy rules in both single-label and multi-label settings, this paper proposes two enhanced methods: FLEL-SL-KNN (for single-label learning) and FLEL-ML-KNN (for multi-label learning) based on traditional KKN-based frameworks to improve learning effectiveness. Comprehensive experiments demonstrate that incorporation of fuzzy labels significantly enhance the performance of traditional label learning methods, highlighting the effectiveness of fuzzy labeling in label learning.

The remainder of this paper is organized as follows: Section 2 reviews existing label learning methods and their limitations. Section 3 introduces the fuzzy label concept. Section 4 presents a fuzzy label generation method. Section 5 proposes fuzzy label-enhanced learning algorithms. Section 6 reports experimental results, and Section 7 concludes the

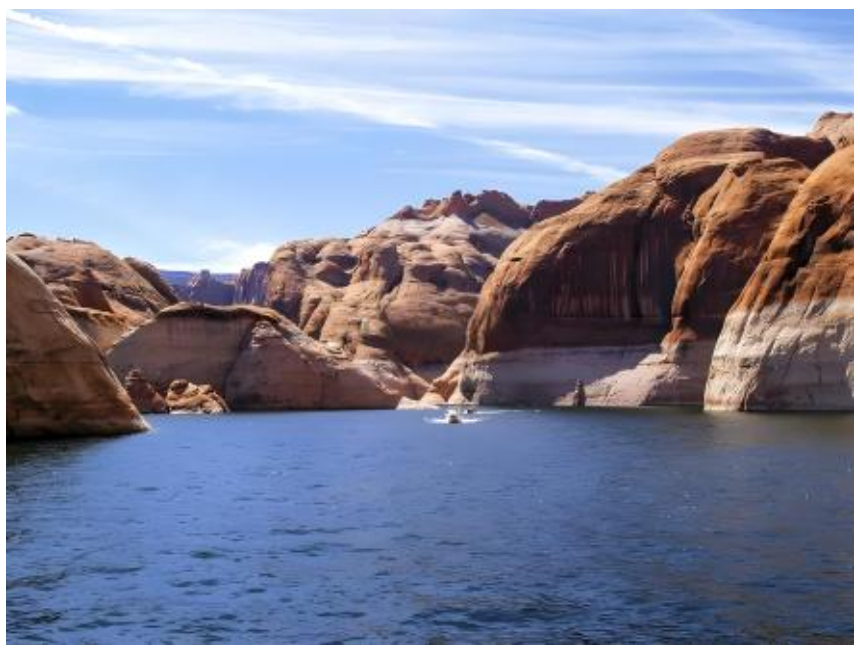

(a) Image Object

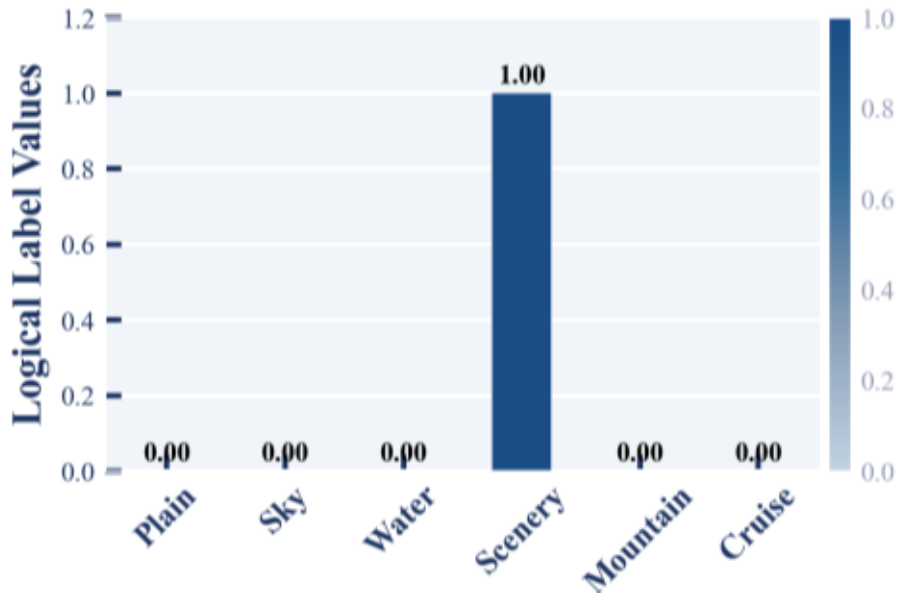

(b.1) Logical Labels (single label)

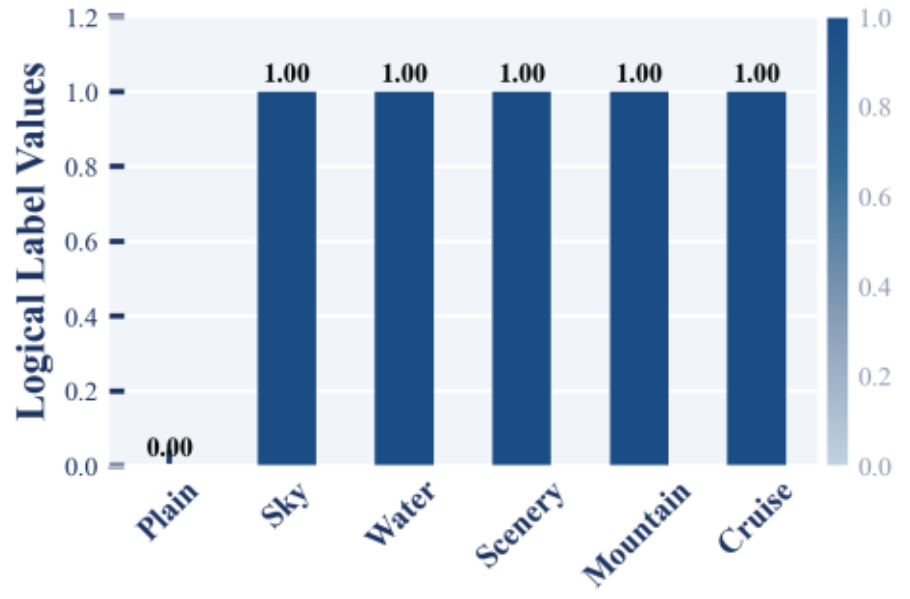

(b.2) Logical Labels (multi-label)

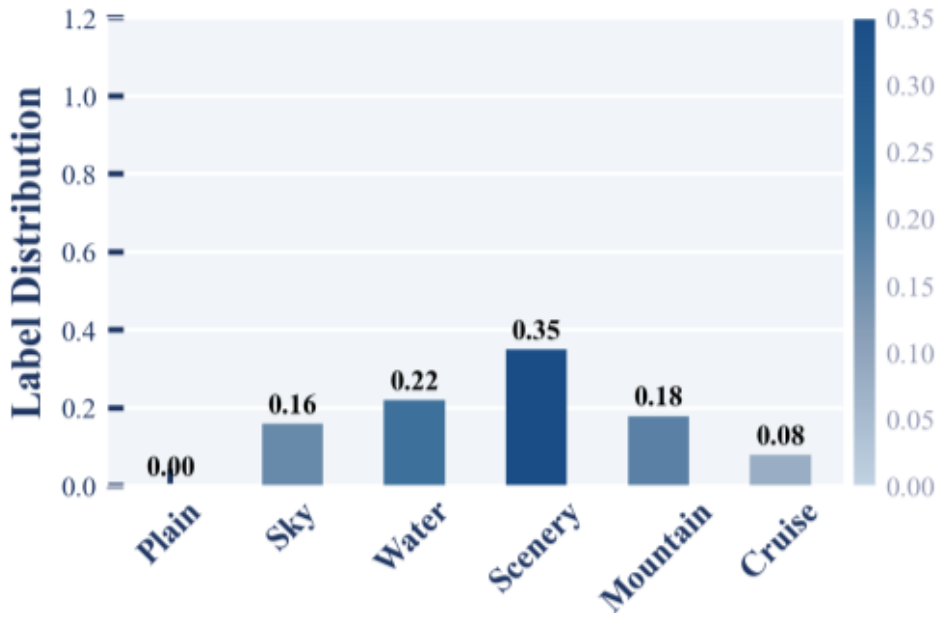

(c) Label Distribution

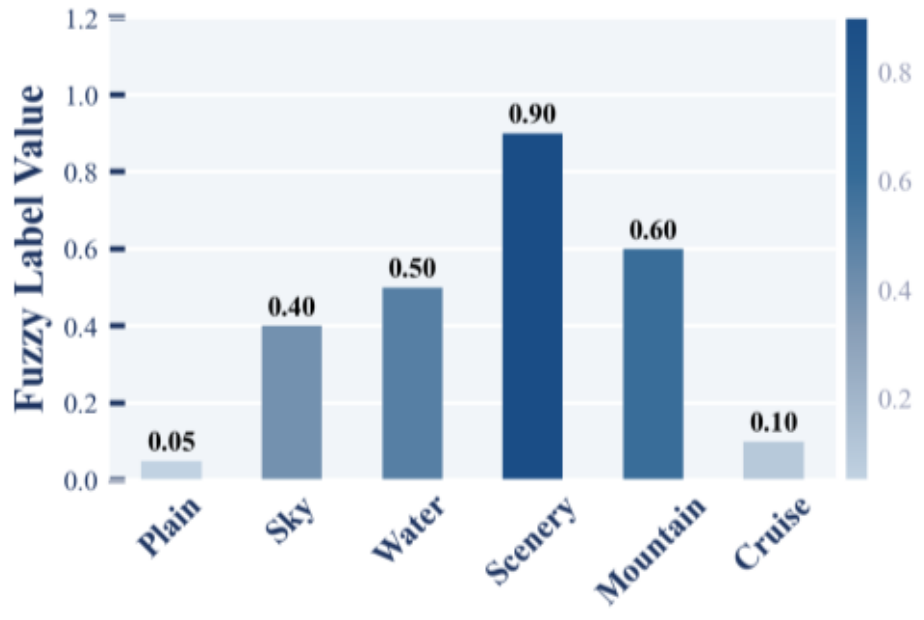

(d) Fuzzy Labels

Fig. 1. The figure illustrates different label representation methods for a natural scene image. Specifically, (a) depicts the original natural scene image, (b) The logical labels for each category within the image are visualized separately under single-label and multi-label classification scenarios. (c) shows the label distribution for each category, and (d) presents the category visualization represented by fuzzy labels.

paper and outline future work.

## II. BACKGROUND

### A. Traditional Label Learning Methods

As illustrated in Fig. 1(a), consider this natural scene image object $\boldsymbol{x}_n$ associated with $L$ ( $L=6$ ) potential labels. The six labels include "Plain", "Sky", "Water", "Scenery", "Mountain" and "Cruise ship". Traditional logical labeling methods use binary indicators - "1" or "0" to signify whether an instance belongs to a specific category [17]-[19]. Let $\boldsymbol{x}_n$ denote the instance (i.e., the iamge) and a label vector $\boldsymbol{y}_n=\left[y_n^1, y_n^2, \cdots, y_n^L\right]$ to represent its relationship with the candidate labels. In this context, the logical label vector for instance $\boldsymbol{x}_n$ can be expressed for both single-label classification and multi-label description scenarios as follows:

$$\begin{gathered}\boldsymbol{y}_n=\left[y_n^1, y_n^2, \cdots, y_n^L\right]=[0,0,0,1,0,0] \\ \text{s.t. } y_n^l \in\{0,1\}\end{gathered} \tag{1}$$

$$\begin{gathered}\boldsymbol{y}_n=\left[y_n^1, y_n^2, \cdots, y_n^L\right]=[0,1,1,1,1,1] \\ \text{s.t. } y_n^l \in\{0,1\}\end{gathered} \tag{2}$$

The label vector in Eq. (2) is visualized in Fig. 1(b.2), where it can be seen that all label values, except for "Plain", are assigned deterministically to the remaining five categories. This results in a rigid labeling strategy that eliminates ambiguity and uncertainty at the label level. However, as illustrated in Fig. 1(a), such an approach overlooks the intrinsic ambiguity associated with certain labels. For example, the "Cruise ship" occupies only a small portion of the image and may be easily confused with the surrounding water, while the "Plain" region lacks distinct structural features and can be mistaken for similar categories like "Rock". In contrast, labels such as "Scenery" and "Sky" exhibit clearer visual semantics and are more confidently identifiable. These logical labeling strategies as represented in Eqs. (1) and (2), fails to capture such label-level uncertainty, leading to a loss in label expressiveness. Consequently, multi-label learning models trained on these overly deterministic labels may disregard important semantic ambiguity, ultimately limiting their both representation capacity and generalization performance.

### B. Label Distribution Learning Method

To address the limitations of logical labels, Geng et al. [20]-[23] proposed the concept of label distribution. This method is grounded in the label distribution assumption, which posits that in multi-label scenarios, different labels contribute unequally to the description of an instance. Therefore, the importance of each label should be represented with varying weights. Taking Fig. 1(a). as an example, the relative importance of the labels can be intuitively ranked as: "Scenery," "Water," "Mountain," "Sky," "Cruise ship," and "plain." In this context, "Scenery," "Water," "Mountain," and "Sky" provide more substantial contributions to the semantic content of the image and should thus be assigned higher weights. In contrast, "Cruise ship" is relatively minor role, contributing less to the overall instance representation. As for the "plain" label, its presence in the image is ambiguous, and its weight in the label distribution may reasonably be set to zero.

Building on this idea, the label distribution method quantitatively expresses the contribution of each label to an instance description using continuous real values within the $[0,1]$ range, rather than relying on binary judgments (i.e., 0 or 1). A key feature of this method is the constraint that the sum of all label contributions for a given instance must equal 1. This normalization ensures that the label distribution captures the semantic association strength between labels and instances from a probabilistic perspective, providing a well-defined mathematical foundation. In practice, this continuous representation enables the model to handle ambiguity and uncertainty inherent in real-world data to some extent. For example, the label distribution of the image in Fig. 1(a) can be formulated as shown in Eq. (3), reflecting the varying importance of each label in describing the instance.

$$\begin{aligned}\boldsymbol{y}_n &= \left[y_n^1, y_n^2, \cdots, y_n^L\right] \\ &= [0.0, 0.16, 0.22, 0.35, 0.18, 0.08] \\ \text{s.t. } & y_n^l \in[0,1],\ \sum_{l=1}^L y_n^l=1\end{aligned} \tag{3}$$

The label distribution vector in Eq. (3) is further visualized in Fig. 1(c), illustrating how each label is assigned a value proportional to its contribution to the instance description. Specifically, the label weights are as follows: [Plain: 0, Sky: 0.16, Water: 0.22, Scenery: 0.35, Mountain: 0.18, Cruise ship: 0.08]. This representation effectively captures the varying degrees of relevance among labels, reflecting both ambiguity and uncertainty than logical labels.

Although label distribution techniques achieve precise quantification of label uncertainty, their distribution concept similar to a probability distribution, deviates from the inherent fuzziness and ambiguity found in real-world labels. This discrepancy primarily arises from the completeness assumption, which enforces mutual exclusivity among label values in the distributions. However, such exclusiveness often fails to reflect reality. For example, as shown in Fig. 1(a), the presence of "Plain" is inherently ambiguous and difficult to define precisely. Moreover, labels such as "Mountain" and "Water" are often subsets of the broader "Scenery" label, and a single instance can exhibit high degrees of membership across these labels simultaneously. These overlapping and uncertain label relationships directly conflict with the completeness assumption underlying label distribution methods. Therefore, the development of novel techniques and theoretical framework that more accurately capture real-world label uncertainty remains an important and open research direction.

## III. THE CONCEPT OF FUZZY LABELING

In this section, we adopt a fuzzy theory perspective to develop new techniques and methods that better capture the uncertainty inherent in real-world labelling. In practice, the attribution of an instance to multiple labels often involves complex interrelationships and varying degrees of membership, exhibiting typical ambiguity. As shown in Fig. 1(a), a instance is simultaneously associated with six different labels. Notably, there are clear differences in the degree of attribution across labels. For example, the instance shows a substantially higher degree of membership to “Scenery” compared to “Cruise ship”. Furthermore, the labels themselves exhibit interdependencies; for instance, “Mountain”, “Water” and “Scenery” are positively correlated. At the level of individual labels, the degrees of attribution also reflect significant ambiguity and uncertainty. For example, the instance is more strongly associated with "Mountain" and "Water", while the associations with "Plain" and "Cruise ship" is notably weaker.

The ambiguity and uncertainty in such attribution relationships primarily arise from two aspects: 1) the inherently fuzzy nature of the instance labels themselves, and 2) subjective differences in judgment among annotators. To better characterize this fuzzy attribution property, we introduce independent fuzzy sets for each label and use membership functions to quantify the degree to which an instance is attributed to a given label. In this paper, we refer to this label representation method as the "Fuzzy Label" method.

For a label learning scenario defined in a feature space $\mathcal{X}$ with $L$ labels, we define fuzzy labels and the corresponding fuzzy label space as follows:

***Definition 1: Fuzzy Label.*** *Let* $A^l$ *denote the fuzzy set corresponding to the* $l$th *label of an instance* $\boldsymbol{x}_n$ *defined in the feature space* $\mathcal{X}$. *Its mathematical representation is expressed as:*

$$A^l = \left\{ \left( \boldsymbol{x}_n, \mu_{A^l}\left(\boldsymbol{x}_n\right) \right) \mid \boldsymbol{x}_n \in \mathcal{X} \right\} \tag{4}$$

*where* $\mu_{A^l} : \mathcal{X} \to [0,1]$ *is the membership function used to characterize the membership of instance* $\boldsymbol{x}_n$ *to* $A^l$. *For any instance* $\boldsymbol{x}_n \in \mathcal{X}$, *its membership to the fuzzy set* $A^l$ *can be represented by the membership* $\mu_{A^l}\left(\boldsymbol{x}_n\right)$:

$$\boldsymbol{x}_n \in \mathcal{X}, \text{ and } \mu_{A^l}\left(\boldsymbol{x}_n\right) \in [0,1] \tag{5}$$

*therefore, the fuzzy membership* $\mu_{A^l}\left(\boldsymbol{x}_n\right)$ *can be used to represent the degree of attribution of instance* $\boldsymbol{x}_n$ *to the* $l$th *label, defined as* $u_n^l = \mu_{A^l}\left(\boldsymbol{x}_n\right)$.

***Definition 2: Fuzzy Label Space.*** *Given a label learning problem defined in the feature space* $\mathcal{X}$ *with* $L$ *labels, a fuzzy set* $A^l$ *is defined for each label. The fuzzy label space is represented as* $\mathcal{U} \subseteq [0,1]^L$. *For any instance* $\boldsymbol{x}_n \in \mathcal{X}$, *its corresponding fuzzy label vector is defined in* $\mathcal{U}$ *as:*

$$\boldsymbol{u}_n = \left[ u_n^1, u_n^2 \cdots, u_n^L \right] \in \mathcal{U}, \; u_n^l = \mu_{A^l}\left(\boldsymbol{x}_n\right) \tag{6}$$

*where each label dimension of the fuzzy label vector satisfies* $u_n^l \in [0,1]$, *representing the fuzzy attribution of instance* $\boldsymbol{x}_n$ *with the corresponding label.*

The advantage of fuzzy labels lies in their ability to precisely express the semantic uncertainty of real-world labels using fuzzy theory. This enables a more faithful representation of the complex and overlapping relationships among labels observed in real-world scenarios.

For the image shown in Fig. 1(a), the fuzzy label vector of instance $\boldsymbol{x}_n$ can be expressed as shown in Eq. (7).

$$\begin{aligned} \boldsymbol{y}_n &= \left[ y_n^1, y_n^2, \cdots, y_n^L \right] \\ &= [0.05, 0.40, 0.50, 0.90, 0.60, 0.10] \\ \text{s.t. } & y_n^l \in [0,1] \end{aligned} \tag{7}$$

The fuzzy label vector in Eq. (7) is also visualized in Fig. 1(d). In this figure, fuzzy labels employ precise membership values to describe the degree of attribution between the instance and each label. Unlike traditional label distributions, fuzzy labels are not constrained by the completeness assumption, which allows an instance to exhibit high degrees of attribution to multiple labels simultaneously.

The fuzzy labeling technique in this paper is designed to address the limitations of traditional logical labels in representing uncertainty. Compared to label distribution techniques, fuzzy labeling offers greater adaptability across different scenarios. To further elucidate the connections and differences between these two methods, a systematic comparison and analysis is provided from the perspectives of single-label and multi-label classification *in Part A of the Supplemental Materials.*

## IV. FUZZY LABEL GENERATION

In practical applications, accurately quantifying the degree of each label often incurs high annotation costs [24]-[26]. Based on this, this section explores effective methods for reconstructing fuzzy label information based on existing logical labels and instance feature data. We define this task as Fuzzy Label Generation (FL-Gen).

### *A. Theoretical Foundations of Fuzzy Label Generation*

Given a $D$-dimensional feature space $\mathcal{X} \subseteq \mathcal{R}^D$ and an $L$-dimensional logical label space $\mathcal{Y} \subseteq \{0,1\}^L$, where each instance $\boldsymbol{x}_n \in \mathcal{X}$ is annotated with $L$ labels, there exists a corresponding logical label vector $\boldsymbol{y}_n = \left[ y_n^1, y_n^2, \cdots, y_n^L \right] \in \mathcal{Y}$ in the logical space $\mathcal{Y}$. Here, $y_n^l \in \{0,1\}$ represents the logical label of $\boldsymbol{x}_n$ for the $l$th label.

Given a dataset $\boldsymbol{S} = \left\{ \left( \boldsymbol{x}_n, \boldsymbol{y}_n \right) \right\}_{n=1}^N$ containing $N$ instances, where $\boldsymbol{x}_n \in \mathcal{X}$ represents the $n$th instance, and $\boldsymbol{y}_n \in \mathcal{Y}$ denotes the corresponding logical label vector for that instance. The goal of FL-Gen is to learn the fuzzy label space $\mathcal{U}$ from the dataset $\boldsymbol{S}$. This enables the assignment of a fuzzy label in $\mathcal{U}$ for each instance in $\mathcal{X}$. A critical theoretical foundation for the task of fuzzy label generation is the smoothness assumption from label propagation theory, which can be formally stated as the following theorem [27].

***Theorem 1: Smoothness Assumption in Label Propagation Theory*** [28]**.** *The label function* $f : \mathcal{X} \to \mathcal{U}$ *defined on the feature space* $\mathcal{X}$ *(mapped to the fuzzy label space* $\mathcal{U}$*) exhibits local smoothness on the data manifold* $M \subset \mathcal{X}$. *Specifically, if two instances* $\boldsymbol{x}_i, \boldsymbol{x}_j \in M$ *are similar in the feature space, their corresponding labels* $\boldsymbol{u}_i$ *and* $\boldsymbol{u}_j$ *should also be similar. Let the similarity on the data manifold be measured by a kernel function* $k : \mathcal{X} \times \mathcal{X} \to \mathbb{R}_+$. *The smoothness assumption requires that the label function satisfies the following constraint:*

$$\min_f \sum_{ij} k\left(\boldsymbol{x}_i, \boldsymbol{x}_j\right) \left\| f\left(\boldsymbol{x}_i\right) - f\left(\boldsymbol{x}_j\right) \right\|^2 + \lambda \cdot \Omega(f) \tag{8}$$

*where* $k\left(\boldsymbol{x}_i, \boldsymbol{x}_j\right)$ *represents the similarity weight between instances* $\boldsymbol{x}_i$ *and* $\boldsymbol{x}_j$; $\Omega(f)$ *is the regularization term used to control the complexity of the function; and* $\lambda > 0$ *is the balancing parameter.*

Based on this smoothness assumption, fuzzy label generation aims to leverage the interrelationships between instances in the feature space $\mathcal{X}$ and the logical label space $\mathcal{Y}$ within the dataset $\boldsymbol{S}$. It seeks to fit a fuzzy label space $\mathcal{U}$ defined on $\mathcal{X}$, along with the corresponding membership functions $u_n^l = \mu_{A^l}(\boldsymbol{x}_n)$ for each label within it. The fuzzy label generation process can be formally defined as follows.

***Definition 3: FL-Gen.*** *Given a dataset* $\boldsymbol{S} = \{(\boldsymbol{x}_n, \boldsymbol{y}_n)\}_{n=1}^{N}$, *where* $\boldsymbol{x}_i \in \mathcal{X}$ *represents the feature vector of the* $ith$ *instance in the feature space* $\mathcal{X}$, *and* $\boldsymbol{y}_i \in \mathcal{Y}$ *represents the logical label vector of the* $ith$ *instance in the logical label space* $\mathcal{Y}$.

*The goal of fuzzy label generation is to fit the following mapping:*

$$f : \mathcal{X} \times \mathcal{Y} \rightarrow \mathcal{U} \tag{9}$$

$$f = \{ \mu_{A^1}(\boldsymbol{x}_n), \mu_{A^2}(\boldsymbol{x}_n), \cdots, \mu_{A^L}(\boldsymbol{x}_n) \} \tag{10}$$

*where* $\mathcal{U} \subseteq [0,1]^K$ *represents the fuzzy label space,* $f$ *is a set of membership functions, and* $\mu_{A^l}(\boldsymbol{x}_n)$ *is the membership function corresponding to the* $lth$ *label. For any instance* $\boldsymbol{x}_i$, *its corresponding fuzzy label vector* $\boldsymbol{u}_i = f(\boldsymbol{x}_i, \boldsymbol{y}_i)$ *satisfies:*

$$\boldsymbol{u}_i = (u_i^1, u_i^2 \cdots, u_i^L), u_n^l = \mu_{A^l}(\boldsymbol{x}_n) \in [0,1] \tag{11}$$

*where* $u_i^l$ *represents the membership degree of instance* $\boldsymbol{x}_i$ *to the* $lth$ *label.*

*B. Fuzzy Label Generation Method Based on Iterative Label Propagation.*

Given a dataset $\boldsymbol{S} = \{(\boldsymbol{x}_n, \boldsymbol{y}_n)\}_{n=1}^{N}$, where $\boldsymbol{x}_n \in \mathcal{X}$ represents its $nth$ instance and $\boldsymbol{y}_n \in \mathcal{Y}$ denotes the corresponding logical label vector. Based on the definitions in the previous section, this section proposes a fuzzy label generation method, Fuzzy Label Generation using Label Propagation (FL-Gen-LP), which is grounded in the smoothness assumption and the spatial clustering assumption, to generate fuzzy labels for the instances in dataset $\boldsymbol{S}$.

Based on the smoothness assumption [29], [30], FL-Gen-LP begins by constructing a fully connected graph $\mathcal{G} = (V, E, \boldsymbol{W})$ to represent the relationships among instances in the dataset. The vertex set $V$ corresponds to all instances in $\boldsymbol{S}$, while the edge set $E$ represents the connections between each instance and its nearest neighbors. A weight matrix $\boldsymbol{W} = [w_{ij}]_{N \times N}$ is introduced to quantify the similarity between instances, where $w_{ij}$ denotes the similarity between instance $\boldsymbol{x}_i$ and instance $\boldsymbol{x}_j$.

To measure the similarity between instances, FL-Gen-LP defines a similarity function, as shown in the Eq. (12), to evaluate the spatial similarity between any two instances. As the function indicates, the closer two instances are in the feature space, the higher their corresponding similarity score.

$$f_{close}(\boldsymbol{x}_i, \boldsymbol{x}_j) = \begin{cases} \exp(-\frac{\|\boldsymbol{x}_i - \boldsymbol{x}_j\|_2^2}{2\sigma^2}) & i \neq j \\ 0 & i = j \end{cases} \tag{12}$$

Additionally, to achieve a more precise measurement of similarity, FL-Gen-LP incorporates the spatial clustering assumption [31] alongside spatial similarity. This assumption posits that if two instances in the feature space belong to the same cluster, they are highly likely to share the same or similar labels.

***Theorem 2: Spatial Clustering Assumption*** [31]. *Assume that the dataset* $\boldsymbol{S}$ *in the feature space* $\mathcal{X}$ *can be divided into* $K$ *clusters* $\{C_k\}_{k=1}^{K}$, *and the labeling function* $f : \mathcal{X} \rightarrow \mathcal{U}$ *map instances to the label space. If it satisfies:*

$$\forall k \in \{1, \ldots, K\},\ \forall \boldsymbol{x}_i, \boldsymbol{x}_j \in C_k, \quad \|f(\boldsymbol{x}_i) - f(\boldsymbol{x}_j)\| \leq \eta\ (\eta \rightarrow 0_+) \tag{13}$$

*then* $\boldsymbol{S}$ *is said to satisfy the spatial clustering assumption.*

Considering the spatial clustering assumption, the dataset $\boldsymbol{S}$ is clustered into $K$ clusters using the Fuzzy C-Means (FCM) algorithm, resulting in a membership matrix $\boldsymbol{M} = [m_n^k]_{N \times K}$. Here, $m_n^k$ denotes the membership degree of instance $\boldsymbol{x}_n$ to the $kth$ cluster. For each instance $\boldsymbol{x}_i$, the index $k_{\max}$ of the cluster to which it has the highest membership degree is determined as follows:

$$k_{i,\max} = \arg\max_k \{ m_i^k \mid k = 1, 2, \cdots, K \} \tag{14}$$

Subsequently, we design a similarity measurement function that integrates the spatial smoothness assumption and the spatial clustering assumption:

$$w_{ij}^{New} = f_{close}(\boldsymbol{x}_i, \boldsymbol{x}_j) \times m_j^{k_{i,\max}} \tag{15}$$

In Eq. (15), $m_j^{k_{i,\max}}$ allows instances within the same cluster to achieve higher similarity when calculating the similarity between instance $\boldsymbol{x}_i$ and other instances, thereby adhering to the spatial clustering assumption.

Based on Eq. (15), the fully connected graph of the instance set is reconstructed, and the weight matrix of the graph $\mathcal{G}$ can be redefined as $\boldsymbol{W}^{New} = [w_{ij}^{New}]_{N \times N}$. After obtaining the weight matrix $\boldsymbol{W}^{New}$, classical iterative label propagation techniques [31] can be employed to learn the fuzzy labels for the instances. First, the label propagation matrix $\boldsymbol{P} = \widehat{\boldsymbol{A}}^{-\frac{1}{2}} \boldsymbol{W} \widehat{\boldsymbol{A}}^{-\frac{1}{2}} \in \mathbb{R}^{N \times L}$ is computed using the weight matrix $\boldsymbol{W}^{New}$. Here, $\widehat{\boldsymbol{A}} = \mathrm{diag}[\widehat{a}_1, \widehat{a}_2 \cdots, \widehat{a}_N]$ is a diagonal matrix where $\widehat{a}_i = \sum_{j=1}^{N} w_{ij}$. Additionally, a non-negative matrix $\boldsymbol{U} = [u_{nl}]_{N \times L}$ is defined to represent the fuzzy label matrix to be learned, which is initialized as $\boldsymbol{U}^{(0)} = [y_{nl}]_{N \times L}$ before the iteration begins. The optimal solution for $\boldsymbol{U}$ can then be obtained through an iterative approach. During the $tth$ iteration, the update strategy for $\boldsymbol{U}$ in the label propagation process is expressed as:

$$\boldsymbol{U}^{(t)} = \alpha \boldsymbol{P} \boldsymbol{U}^{(t-1)} + (1 - \alpha) \boldsymbol{Y} \tag{16}$$

In this update strategy, the first term $\alpha \boldsymbol{P} \boldsymbol{U}^{(t-1)}$ is used to propagate relational information among instances, while the second term $(1-\alpha)\boldsymbol{Y}$ preserves and inherits the existing logical label information. The relative importance of these two terms is balanced by a controlling parameter $\alpha \in (0,1)$, which adjusts the influence of propagation versus inheritance in the final label assignment. The detailed algorithmic process of the fuzzy label generation algorithm, FL-Gen-LP, is described in Algorithm I *in Part B of the Supplemental Materials.*

## V. FUZZY LABEL ENHANCEMENT LEARNING

In order to verify the usefulness and importance of fuzzy labels in label learning, this section explores fuzzy label-based enhancement learning using two classical label learning algorithms: 1) K-Nearest Neighbor (KNN) for single-label learning and 2) ML-KNN for multi-label learning. By incorporating fuzzy labels, we propose two enhanced variants: Fuzzy Single-Label Enhancement Learning based KNN (FLEL-SL-KNN) and Fuzzy Multi-Label Enhancement Learning based ML-KNN (FLEL-ML-KNN). The former improves single-label classification through a fuzzified nearest neighbor voting mechanism, while the latter extends to multi-label scenarios by leveraging quantitative fuzziness to improve modeling of label relevance.

### A. FLEL-SL-KNN

Traditional KNN predicts the class of a test instance based on its nearest neighbors and their crisp labels. These simple labels cannot adequately capture the underlying uncertainty present in the data. To address this limitation, FLEL-SL-KNN incorporates fuzzy labels during prediction, providing richer semantic information about label uncertainty, enabling KNN to perform better in complex and ambiguous situations.

Specifically, given a single-label training dataset $\boldsymbol{S}=\left\{\left(\boldsymbol{x}_n,\boldsymbol{y}_n\right)\right\}_{n=1}^{N}$, where $\boldsymbol{x}_n\in\mathcal{X}\subseteq\mathcal{R}^D$ is the $nth$ instance, $\boldsymbol{y}_n\in\mathcal{Y}\subseteq\{0,1\}^C$ is the corresponding logical label vector of the instance, $D$ is the feature dimension, and $C$ is the number of classes. The fuzzy labels $\boldsymbol{u}_k=\left[u_k^1,u_k^2,\cdots,u_k^C\right]\in\mathbb{R}^{1\times C}$ for each instance in the dataset $S$ are obtained using the FL-Gen-LP algorithm.

In the inference process of FLEL-SL-KNN, for each test instance $\boldsymbol{x}_{test}\in\mathcal{X}$, the distances between it and all training instances $\boldsymbol{x}_{train}\in\mathcal{S}$ are first computed. Based on these distances, the $K$ nearest neighbors are selected from the training set to form the nearest neighbor set $\mathcal{N}_K\left(\boldsymbol{x}_{test}\right)$. FLEL-SL-KNN employs the Euclidean distance, as defined in Eq. (17), as the distance metric.

$$\mathcal{D}\left(\boldsymbol{x}_{test},\boldsymbol{x}_{train}\right)=\sqrt{\sum_{d=1}^{D}\boldsymbol{x}_{test}^d-\boldsymbol{x}_{train}^d} \tag{17}$$

Subsequently, the fuzzy label vector of the test instance $\boldsymbol{x}_{test}$ is computed by performing a weighted aggregation of the fuzzy label vectors of its nearest neighbors. The weights are determined based on the similarity between the test instance and each of its neighboring instances. Specifically, the fuzzy label vector of the test instance is calculated as follows:

$$\boldsymbol{u}_{test}=\frac{\sum_{\boldsymbol{x}_k\in\mathcal{N}_K\left(\boldsymbol{x}_{test}\right)}w_{test,k}\cdot\boldsymbol{u}_k}{\sum_{\boldsymbol{x}_k\in\mathcal{N}_K\left(\boldsymbol{x}_{test}\right)}w_{test,k}} \tag{18}$$

where $\boldsymbol{u}_k$ represents the fuzzy label vector of the neighboring instance $\boldsymbol{x}_k$, and $w_{test,k}$ denotes the similarity between the test instance $\boldsymbol{x}_{test}$ and the neighboring instance $\boldsymbol{x}_k$, which can be calculated using Equation (19).

$$w_{test,k}=\frac{1}{\mathcal{D}\left(\boldsymbol{x}_{test},\boldsymbol{x}_k\right)+\varepsilon} \tag{19}$$

where $\varepsilon$ is a small constant introduced to prevent division by zero errors.

After obtaining the fuzzy label of $\boldsymbol{x}_{test}$, a maximum value method can be used to define the conversion function that maps the fuzzy labels to logical labels. This process transforms the fuzzy label back into the final logical label $\boldsymbol{y}_{test}=\left[y_{test}^1,y_{test}^2,\cdots,y_{test}^C\right]\in\mathbb{R}^{1\times C}$ for single-label classification. The label conversion function used in FLEL-SL-KNN can be expressed as:

$$y_{test}^c=\begin{cases}1, \text{ if } c=\arg\max_{c\in\{1,2,\cdots,C\}}\left(u_{test}^c\right)\\ 0, \text{ otherwise}\end{cases} \tag{20}$$

where $\arg\max$ denotes the operation that returns the category $c$ corresponding to the element in the fuzzy label vector $\boldsymbol{u}_{test}$ with the maximum value.

Based on the above analysis, the prediction process of the FLEL-SL-KNN classifier is provided in Algorithm II *in Part B of the Supplemental Materials.*

### B. FLEL-ML-KNN

Given a multi-label training dataset $\boldsymbol{S}=\left\{\left(\boldsymbol{x}_n,\boldsymbol{y}_n\right)\right\}_{n=1}^{N}$, where $\boldsymbol{x}_n\in\mathcal{X}\subseteq\mathcal{R}^D$ represents the $nth$ instance with $D$ features, and $\boldsymbol{y}_n\in\mathcal{Y}\subseteq\{0,1\}^L$ denotes its corresponding logical label vector with $L$ labels. Similarly, the FL-Gen-LP algorithm is employed to obtain the fuzzy labels $\boldsymbol{u}_k=\left[u_k^1,u_k^2,\cdots,u_k^C\right]\in\mathbb{R}^{1\times L}$ for each instance in the dataset $\boldsymbol{S}$.

During the inference process of FLEL-ML-KNN, for each test instance $\boldsymbol{x}_{test}\in\mathcal{X}$, the distance between $\boldsymbol{x}_{test}$ and all training instances $\boldsymbol{x}_{train}\in\mathcal{S}$ are calculated. Based on these distances, the $K$-nearest neighbors are identified, forming a nearest neighbor set $\mathcal{N}_K\left(\boldsymbol{x}_{test}\right)$ from the training dataset.

After obtaining the nearest neighbor set, FLEL-ML-KNN first calculates the prior probability $P_{prior}(l)$ for each label $l\in\{1,2,\cdots,L\}$ in the fuzzy labels of the training set. The calculation method is given by Eq. (21).

$$P_{prior}(l)=\frac{\mathcal{C}_{global}(l)+S_{smooth}}{N+2\times S_{smooth}} \tag{21}$$

$$\mathcal{C}_{global}(l)=\sum\nolimits_{\boldsymbol{x}_{train}\in\mathcal{S}}u_n^l,\ l\in\{1,2,\cdots,L\} \tag{22}$$

where $N$ represents the total number of training instances in $S$, and $\mathcal{C}_{global}(l)$ denotes the sum of the fuzzy label values for label $l$ across all training instances in $S$. $S_{smooth}$ is a smoothing constant introduced to prevent division by zero during computation.

Subsequently, fuzzy labels are employed to compute the conditional probability $P_{cond_l}$ and $P_{condN_l}$ for the test instance $\boldsymbol{x}_{test}$ within its nearest neighbor set. These represent the likelihoods of the instance belonging to and not belonging to class $l$, respectively. A threshold $\sigma$ is defined for each label $l$. If a fuzzy label value of an instance for label $l$ exceeds $\sigma$, the instance is considered to belong to label $l$; otherwise, it is considered not to belong to label $l$. Based on this, the frequency of instances within the $K$- nearest neighbors of instance $\boldsymbol{x}_{test}$ that belong to label $l$, as well as the frequency of instances not belonging to label $l$, can be computed as shown in Eqs. (23) and (24).

$$count_l\left(\boldsymbol{x}_{test}\right)=\sum_{\boldsymbol{x}_k\in\mathcal{N}_k\left(\boldsymbol{x}_{test}\right)}\mathbb{I}\left(u_n^l\le\sigma\right) \tag{23}$$

$$countN_l\left(\boldsymbol{x}_{test}\right)=\sum_{\boldsymbol{x}_k\in\mathcal{N}_k\left(\boldsymbol{x}_{test}\right)}\mathbb{I}\left(u_n^l\le\sigma\right) \tag{24}$$

where $\mathbb{I}(*)$ denotes the indicator function, which equals 1 if $u_n^l>\sigma$; otherwise, it equals 0. Subsequently, based on the distribution of fuzzy labels within the nearest neighbor set $\mathcal{N}_K\left(\boldsymbol{x}_{test}\right)$ of the test instance $\boldsymbol{x}_{test}$, the conditional probability $P_{cond_l}$ and $P_{condN_l}$ for belonging to and not belonging to the category, respectively, can be computed.

$$P_{cond_l}\approx\frac{S_{smooth}+count_l\left(\boldsymbol{x}_{test}\right)}{S_{smooth}\times\left(\mathrm{K}+1\right)+\sum_{l\in\{1,2,\cdots,L\}}count_l\left(\boldsymbol{x}_{test}\right)} \tag{25}$$

$$P_{condN_l}\approx\frac{S_{smooth}+countN_l\left(\boldsymbol{x}_{test}\right)}{S_{smooth}\times\left(K+1\right)+\sum_{l\in\{1,2,\cdots,L\}}countN_l\left(\boldsymbol{x}_{test}\right)} \tag{26}$$

In the posterior probability calculation, the use of fuzzy labels combined with a threshold enables more effective handling of label uncertainty. This approach facilitates a clearer and more reasonable separation of classification boundaries for each label, leading to a more accurate estimation of the underlying fuzzy label distribution. Ultimately, for each test instance $\boldsymbol{x}_{test}$, the fuzzy label for category $l$ can be calculated using Bayes' theorem:

$$P\left(l\mid\boldsymbol{x}_{test}\right)=\frac{P_{prior}\left(l\right)\times P_{cond_l}}{P_{prior}\left(l\right)\times count_l\left(\boldsymbol{x}_n\right)+P_{prior}\left(\neg l\right)\times P_{condN_l}} \tag{27}$$

where $P_{prior}\left(\neg l\right)=1-P_{prior}\left(l\right)$ represents the inverse prior probability for category $l$.

After obtaining the fuzzy labels, the fuzzy labels are re-transformed into the final logical label $\boldsymbol{y}_{test}=\left[y_{test}^1,y_{test}^2,\cdots,y_{test}^L\right]\in\mathbb{R}^{1\times L}$ using the threshold function as shown in Eq. (28).

$$y_{test}^l=\begin{cases}1, & \text{if } P\left(l\mid\boldsymbol{x}_{test}\right)\ge\sigma\\0, & \text{if } P\left(l\mid\boldsymbol{x}_{test}\right)<\sigma\end{cases} \tag{28}$$

Based on the above analysis, the training and inference process of the fuzzy label enhancement-based multi-label classifier, FLEL-ML-KNN, is detailed in Algorithm III *in Part B of the Supplemental Materials.*

## VI. EXPERIMENTAL ANALYSIS

TABLE I
SINGLE-LABEL DATASET DESCRIPTION

| **Data set** | **#Instance** | **#Feature** | **#Class** |
|---|---|---|---|
| Artificial(Ar.) | 1500 | 5 | 3 |
| divorce | 169 | 54 | 2 |
| heart | 303 | 13 | 2 |
| breast cancer | 569 | 30 | 2 |
| wine | 178 | 13 | 3 |
| waveform | 5000 | 22 | 3 |

### *A. Single-Label Classification Experiments*

In this section, we evaluate the proposed FLEL-SL-KNN on multiple single-label datasets to verify the effectiveness of introducing fuzzy labels in enhancing label learning performance for single-label learning tasks.

#### *a) Single-Label Datasets*

This study uses five real-world single-label datasets and one artificially designed dataset for experimental evaluation. Table I summarizes the key characteristics of each dataset. The real datasets, sourced from various domains, are publicly available from UCI repository [1]. The artificial dataset is specifically designed to intuitively demonstrate the effectiveness of the proposed fuzzy label generation method, with its construction details provided in Part C of the supplementary material.

#### *b) Experiment Settings*

The proposed method FLEL-SL-KNN is developed by introducing fuzzy labels based on the traditional KNN method. Experiments compare FLEL-SL-KNN and KNN on several datasets using three common metrics: Accuracy, F1-score, and ROC-AUC [32], [33]. Accuracy and F1-score measure overall correctness and balance, while ROC-AUC evaluates the ability to distinguish positive and negative classes at different thresholds. Detailed metric definitions are *in Part D of the supplementary material.*

In all experiments, five-fold cross-validation is used to evaluate the performance of each model. The average value of each evaluation metric is recorded to analyze the model's behavior. For both algorithms, the range of nearest neighbor parameters is set as $K=\{1,3,5,7,9,11,13\}$, which helps analyze how performance changes when using different numbers of neighbors.

#### *c) Experimental Results and Analysis*

#### *1. Experimental comparison of artificial datasets*

To validate the effectiveness of the proposed FL-Gen-LP and its integration into FLEL-SL-KNN, we designed three sets of experiments for systematic evaluation. First, the relevant concepts in the construction of synthetic data are defined as follows: in the single-label synthetic dataset, the artificially designed fuzzy labels are referred to as *true fuzzy labels*; the logical labels obtained by applying a transformation function with a threshold of 0.5 to the true fuzzy labels are referred to as *true logical labels*; and the fuzzy labels generated by the fuzzy label generation algorithm based on the true logical labels are referred to as *generated fuzzy labels*.

Based on the above definition, the experiment is divided into the following three parts:

(1) Exp 1 classifies test instances using FLEL-SL-KNN based on the true logical labels of the training instances. In this case, FLEL-SL-KNN degenerates into the traditional soft voting-based KNN.

(2) Exp 2 classifies test instances using the FLEL-SL-KNN model based on the true fuzzy labels of the training instances.

(3) Exp 3 classifies test instances using the FLEL-SL-KNN model based on the generated fuzzy labels of the training instances.

[1] https://archive.ics.uci.edu

TABLE II
PERFORMANCE COMPARISON OF THE PROPOSED FLEL-SL-KNN UNDER THREE DIFFERENT SETTINGS ON THE SYNTHETIC DATASET

| Dataset | Experiment | Metric | | |
|---|---|---|---|---|
| | | Accuracy | F1 | AUC |
| D1 (Training set and test set both using the true logical label) | **exp1** | 0.9183 | 0.9350 | 0.9482 |
| D2 (Training set using the true fuzzy label and test set using the true logical label) | **exp2** | **0.9417** | **0.9416** | 0.9564 |
| D3 (Training set using the generated fuzzy label and test set using the true logical label) | **exp3** | 0.9410 | 0.9412 | **0.9565** |

TABLE III
PERFORMANCE COMPARISON BETWEEN THE PROPOSED FLEL-SL-KNN AND ON REAL DATASETS

| Dataset | Method | Metric | | |
|---|---|---|---|---|
| | | Accuracy | F1 | AUC |
| **divorce** | KNN | 0.9567 | 0.9664 | **0.9769** |
| | FSL_KNN | **0.9779** | **0.9776** | 0.9754 |
| **breast_cancer** | KNN | 0.9145 | 0.9201 | 0.9044 |
| | FSL_KNN | **0.9583** | **0.9549** | **0.9544** |
| **heart** | KNN | 0.7541 | 0.7023 | 0.7705 |
| | FSL_KNN | **0.7984** | **0.7948** | **0.7932** |
| **wine** | KNN | 0.8951 | 0.9224 | 0.9431 |
| | FSL_KNN | **0.9580** | **0.9599** | **0.9712** |
| **waveform** | KNN | 0.8167 | 0.7952 | 0.8266 |
| | FSL_KNN | **0.8297** | **0.8253** | **0.8727** |

The experimental results are shown in Table II. Based on the results in Table II, the following analysis can be made:

(1) Compared with exp 1, exp 2 shows better results on all metrics. This suggests that the FLEL-SL-KNN classifier using true fuzzy labels can handle label uncertainty more effectively. It outperforms the traditional KNN model in most evaluation metrics, especially in recall and F1-score. This shows that the fuzzy classifier can better use the information in fuzzy labels to improve classification.

(2) Compared with exp 1, exp 3 also shows better performance on all evaluation metrics. This confirms that the fuzzy label generation algorithm helps improve classification results. Although the generated fuzzy labels are not exactly the same as the true ones, they still capture useful fuzzy information and boost the classifier's performance. This shows that even without true fuzzy labels, the generated ones can still improve model performance.

(3) The results of exp 2 and 3 are similar, showing that both generated and true fuzzy labels help improve single-label models. This proves that the proposed fuzzy label generation method works well. Even when using generated fuzzy labels instead of true ones, the fuzzy-label-enhanced classifier still performs strongly. This shows that the FL-Gen-LP method is both useful and reliable, and the fuzzy label generation plays an important role in fuzzy-label-based learning.

*2. Experimental Comparison over Real Datasets*

Table III shows that FLEL-SL-KNN significantly outperforms traditional KNN on five real-world datasets. Visualizations are provided *in Part F of the Supplemental Materials*, and the time complexity analysis of FLEL-SL-KNN is presented *in Part G of the Supplemental Materials.* By capturing complex relationships between instances and labels, fuzzy labels address the limitations of traditional hard labeling and more accurately reflect the latent ambiguity in data. This design not only enhances the informational capacity of labels but also improves the model's adaptability to complex data distributions and uncertainties. Secondly, unlike the hard voting mechanism of traditional KNN, FLEL-SL-KNN adopts a distance-based weighted fusion strategy, where closer neighboring instances are assigned higher weights in the prediction process. This strategy improves predictions by reducing errors from complex data and enhancing robustness and accuracy in handling ambiguity and subtle differences.

In summary, FLEL-SL-KNN seamlessly combines fuzzy labels with a weighted fusion strategy, significantly improving the model's classification capability and stability. In terms of classification accuracy and other evaluation metrics, this method always performs better than traditional KNN, showing its benefits and practical use in handling unclear and uncertain data.

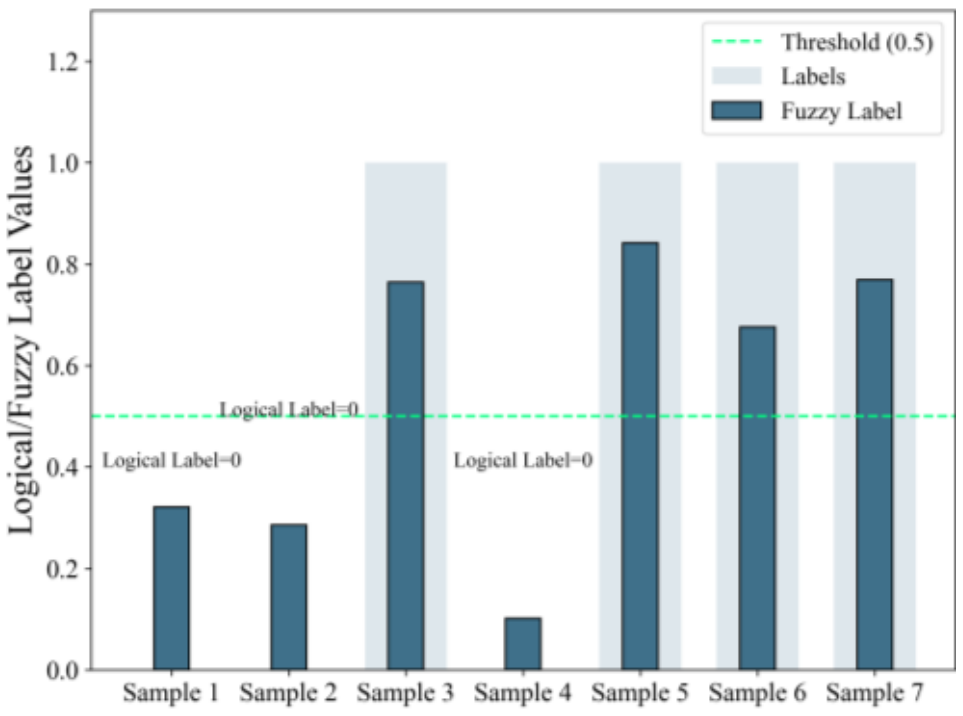

Fig. 2. Comparison of fuzzy labels after logical labeling and fuzzy labeling generation for the first 7 samples under the dataset "heart".

*d) Fuzzy Labels Visualization*

TABLE V
MEAN (SD) OF ALL PERFORMANCE METRICS FOR THE PROPOSED FLEL-ML-KNN UNDER THREE DIFFERENT SETTINGS ON THE SYNTHETIC DATASET.

| Dataset | Experiment | Metric | | | | |
|---|---|---|---|---|---|---|
| | | AP ↑ | HL ↓ | OE ↓ | RL ↓ | CV ↓ |
| **D1 (Training set and test set both using the true logical label)** | **exp 1** | 0.5698 (0.02) | 0.3384 (0.01) | 0.5787 (0.03) | 0.3884 (0.01) | 0.1938 (0.01) |
| **D2 (Training set using the true fuzzy label and test set using the true logical label)** | **exp 2** | 0.6178 (0.02) | 0.2773 (0.00) | 0.4753 (0.03) | 0.3085 (0.01) | 0.1618 (0.02) |
| **D3 (Training set using the generated fuzzy label and test set using the true logical label)** | **exp 3** | **0.6264 (0.02)** | **0.2340 (0.01)** | **0.4647 (0.03)** | **0.2894 (0.02)** | **0.1518 (0.02)** |

To intuitively demonstrate the effectiveness of the fuzzy label generation algorithm FL-Gen-LP, we selected the first seven instances from the "divorce" dataset and conducted a visual comparison between the generated fuzzy labels and the original logical labels. The results are presented in Fig. 2. As shown in the figure, FL-Gen-LP generates fuzzy labels that are closely aligned with the instance features, effectively capturing the latent associations between the instances and their corresponding labels. Moreover, fuzzy labels provide a detailed representation of label uncertainty, which solves the limitations of logical labels in expressing complex relationships and handling boundary ambiguities.

*B. Multi-Labeling Classification Experiments*

In this section, we evaluate the proposed FLEL-ML-KNN on multiple multi-label datasets to verify the enhancement effect of fuzzy labels on label learning performance in multi-label learning tasks.

*a) Multi-Label Dataset*

This study evaluates multi-label learning tasks using nine real-world datasets and one artificially designed dataset. Details are summarized in Table IV. The real-world datasets span diverse domains and are available through MULAN[2]. The artificial dataset is designed to intuitively demonstrate the effectiveness of the generated fuzzy multi-labels, with its construction described *in Part C of the Supplementary Materials.*

TABLE IV
MULTI-LABEL DATASET DESCRIPTION

| Data set | #Instance | #Feature | #Label |
|---|---|---|---|
| Artificial(Ar.) | 2601 | 3 | 3 |
| Flags | 194 | 19 | 7 |
| Cal500 | 502 | 68 | 174 |
| Birds | 645 | 260 | 19 |
| Emotions | 593 | 72 | 6 |
| Genbase | 662 | 1186 | 27 |
| Yeast | 2417 | 103 | 14 |
| Scene | 2407 | 294 | 6 |
| Rcv1(subset1) | 6000 | 944 | 101 |
| Mirflickr | 25000 | 1836 | 159 |

*b) Experiment Settings*

The proposed FLEL-ML-KNN method introduces fuzzy labels based on the traditional ML-KNN. Thus, experiments compare the performance of FLEL-ML-KNN and ML-KNN on several datasets. The experiments use five commonly used classification evaluation metrics [8], [34]: Average Precision (AP), Hamming Loss (HL), One Error (OE), Ranking Loss (RL) and Coverage (CV). Detailed definitions of these metrics are provided *in Part D of the Supplementary Materials.*

All experiments use five-fold cross-validation to evaluate model performance. By recording the mean and standard deviation of each evaluation metric, a detailed analysis of the model's performance is conducted. Regarding parameter settings, the range of the nearest neighbor parameter is $K=\{1, 3, 5, 7, 9, 13\}$, and the smoothing parameter was set to $S_{smoth}=\{0.01, 0.03, 0.05, 0.07, 0.09\}$. This experimental design, combining cross-validation with parameter optimization, ensures the reliability and comprehensiveness of the results.

*c) Experimental Results and Analysis*

*1. Experimental comparison of artificial datasets*

To validate the effectiveness of the proposed FL-Gen-LP and its integration into FLEL-ML-KNN, we first define the relevant concepts in artificial data construction as follows: In the multi-label artificial dataset, the artificially designed fuzzy labels are referred to as *true fuzzy labels*; the logical labels obtained by applying the threshold function in Eq. (28) (with a threshold of 0.5) to the true fuzzy labels are referred to as *true logical labels*; and the fuzzy labels generated by the fuzzy label generation algorithm based on the true logical labels are referred to as *generated fuzzy labels*.

Based on these definitions, the experiments are divided into the following three parts:

(1) Exp 1 classifies test instances using FLEL-ML-KNN based on the true logical labels of the training instances. At this point, FLEL-ML-KNN reduces to the traditional ML-KNN.
(2) Exp 2 classifies test instances using the FLEL-ML-KNN model based on the true fuzzy labels of the training instances.
(3) Exp 3 classifies test instances using the FLEL-ML-KNN model based on the generated fuzzy labels of the training instances.

The experimental results are shown in Table V. Based on the results in Table V, the following analysis can be made:

(1) The results obtained in exp 2 demonstrates a significant improvement in classification performance compared to

[2] http://mulan.sourceforge.net/datasets-mlc.html

the results obtained in exp 1, indicating that incorporating true fuzzy labels effectively enhances the model's ability to represent data uncertainty and capture finer-grained label information.

(2) The results of exp 3 outperform those of exp 1, suggesting that although the generated fuzzy labels are derived from the true logical labels, they still effectively capture the underlying label relationships and uncertainty features in the data, thereby improving classification accuracy.

(3) Finally, both exp 2 and exp 3 achieve significantly better performance than exp 1, with exp 3 slightly surpassing Experiment 2. This outcome is likely due to some noise in the true fuzzy labels, which the fuzzy label generation method reduces, leading to better classification performance. These findings validate the effectiveness and robustness of the fuzzy multi-label generation method in enhancing the performance of multi-label classification tasks.

*2. Experimental comparison of real datasets*

Table VI compares the performance of ML-KNN and FLEL-ML-KNN on nine real-world datasets. Visualization and further analysis are provided in Part F of the Supplemental Materials, while the time complexity analysis of FLEL-ML-KNN is detailed *in Part G of the Supplemental Materials.* The experimental results demonstrate that FLEL-ML-KNN consistently outperforms ML-KNN across all evaluation metrics, further validating the necessity of introducing fuzzy labels. Unlike traditional hard labels, multi-label fuzzy labels are grounded in fuzzy set theory and use membership degrees to quantify the association between instances and multiple labels. This enables the model to more accurately represent the uncertainty inherent in real-world data. Such a flexible labeling approach addresses the limits of conventional hard labels in handling multi-label ambiguity, which allows the transformed fuzzy labels to provide richer information. This improves the ability of model to effectively capture complex multi-label relationships.

Moreover, fuzzy labels play a crucial role in optimizing the K-nearest neighbors algorithm. In FLEL-ML-KNN, by calculating the distances between instances and adjusting the contribution of neighbors based on the conditional probabilities of fuzzy labels, the model evaluates instance similarity more accurately. Compared to ML-KNN, FLEL-ML-KNN not only better leverages information from neighboring instances but also effectively captures the complex correlations between labels, demonstrating superior

TABLE VI
MEAN (SD) OF ALL PERFORMANCE METRICS FOR THE PROPOSED FLEL-ML-KNN AND ML-KNN UNDER THE OPTIMAL PARAMETER SETTINGS.

| Dataset | Method | Metric | | | | |
|---|---|---|---|---|---|---|
| | | **AP↑** | **HL↓** | **OE↓** | **RL↓** | **CV↓** |
| **Flags** | **ML-KNN** | 0.8049 (0.02) | 0.2879 (0.01) | **0.2391 (0.05)** | 0.2215 (0.03) | 0.5407 (0.02) |
| | **FLEL-ML-KNN** | **0.8165 (0.01)** | **0.2563 (0.00)** | 0.2754 (0.03) | **0.2107 (0.01)** | **0.5311 (0.02)** |
| **CAL500** | **ML-KNN** | 0.4868 (0.01) | 0.1370 (0.01) | **0.1156 (0.02)** | 0.1926 (0.00) | 0.7507 (0.01) |
| | **FLEL-ML-KNN** | **0.4964 (0.01)** | **0.1300 (0.01)** | 0.2388 (0.01) | **0.1887 (0.01)** | **0.7421 (0.01)** |
| **Emotions** | **ML-KNN** | 0.7370 (0.01) | 0.2352 (0.00) | 0.3507 (0.01) | 0.2403 (0.00) | 0.3715 (0.01) |
| | **FLEL-ML-KNN** | **0.8018 (0.02)** | **0.1885 (0.02)** | **0.2683 (0.05)** | **0.1612 (0.03)** | **0.2860 (0.03)** |
| **Yeast** | **ML-KNN** | 0.7386 (0.01) | 0.2152 (0.00) | 0.2487 (0.00) | 0.1847 (0.00) | 0.4545 (0.00) |
| | **FLEL-ML-KNN** | **0.7761 (0.01)** | **0.1842 (0.01)** | **0.2317 (0.01)** | **0.1634 (0.01)** | **0.4351 (0.01)** |
| **Medical** | **ML-KNN** | 0.7062 (0.01) | 0.0195 (0.00) | 0.3139 (0.03) | 0.0873 (0.00) | 0.1118 (0.01) |
| | **FLEL-ML-KNN** | **0.7954 (0.01)** | **0.0157 (0.01)** | **0.2671 (0.01)** | **0.0431 (0.01)** | **0.0733 (0.01)** |
| **Birds** | **ML-KNN** | 0.3332 (0.03) | 0.0511 (0.00) | 0.6884 (0.02) | 0.1952 (0.01) | 0.1322 (0.00) |
| | **FLEL-ML-KNN** | **0.3550 (0.01)** | **0.0362 (0.01)** | **0.6515 (0.01)** | **0.1817 (0.01)** | **0.1246 (0.01)** |
| **Scene** | **ML-KNN** | 0.8071 (0.01) | 0.1142 (0.00) | 0.2862 (0.01) | 0.1436 (0.01) | 0.1364 (0.01) |
| | **FLEL-ML-KNN** | **0.8549 (0.01)** | **0.0918 (0.00)** | **0.2368 (0.01)** | **0.0910 (0.00)** | **0.0904 (0.00)** |
| **Mirflickr** | **ML-KNN** | **0.5123 (0.00)** | 0.1593 (0.00) | **0.5637 (0.01)** | 0.2107 (0.00) | 0.4511 (0.00) |
| | **FLEL-ML-KNN** | 0.4201 (0.00) | **0.1406 (0.00)** | 0.5865 (0.01) | **0.3280 (0.01)** | **0.4511 (0.00)** |
| **Rcv1 (subset1)** | **ML-KNN** | 0.4211 (0.00) | 0.0278 (0.00) | 0.5572 (0.01) | 0.1356 (0.01) | 0.2755 (0.00) |
| | **FLEL-ML-KNN** | **0.5026 (0.01)** | **0.0275 (0.00)** | **0.5098 (0.00)** | **0.0901 (0.02)** | **0.1892 (0.00)** |

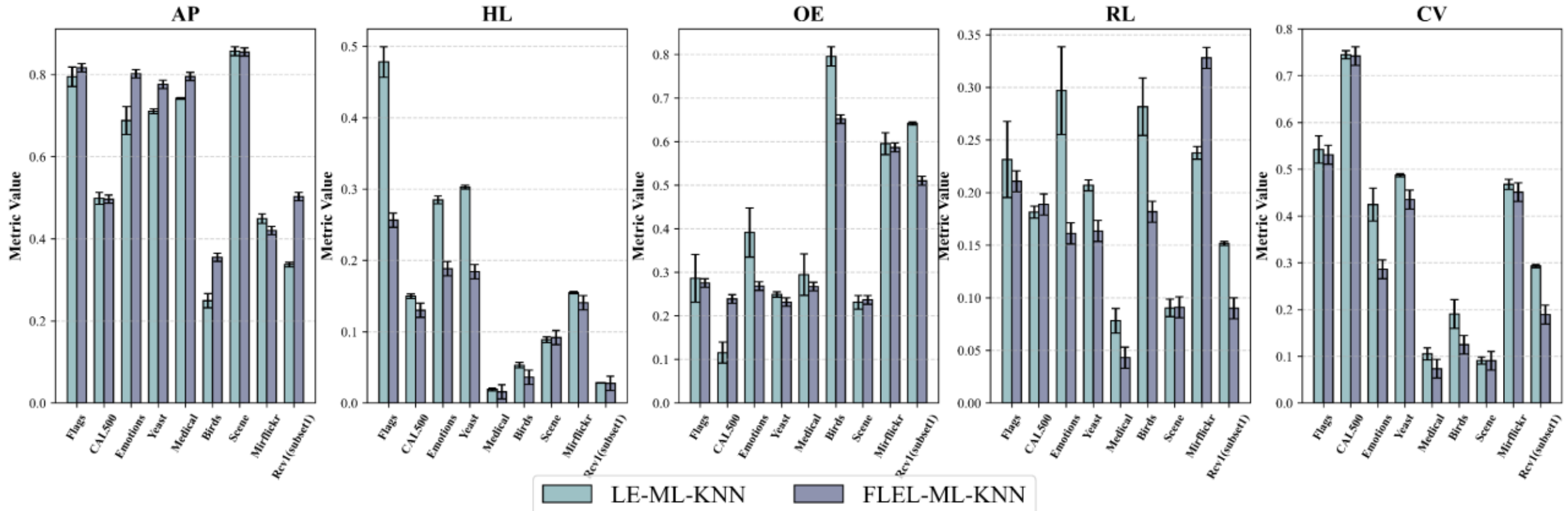

Fig. 3. Performance comparison of LE-ML-KNN and FLEL-ML-KNN on nine multi-label datasets using five evaluation metrics.

classification performance in cases with fuzzy label boundaries or strong multi-label correlations.

The experimental results further demonstrate that FLEL-ML-KNN significantly outperforms ML-KNN in terms of classification accuracy and other key evaluation metrics across nine real-world datasets. This outcome validates the effectiveness of fuzzy labels, indicating that by providing fine-grained membership information for labels, FLEL-ML-KNN can more accurately adjust distance calculations and weight distribution, thereby optimizing classification results.

### d) *Performance Comparison of Different Soft Label Generation Methods*

To evaluate the effectiveness of the proposed fuzzy label generation method (FLEL) against the existing label enhancement method (LE) [23], [28] in terms of latent information mining and model learning capability, this section presents a comparative study conducted on real-world datasets. Specifically, soft labels were generated on nine publicly available multi-label datasets using both methods. These soft labels were then employed to enhance the traditional multi-label classification algorithm ML-KNN, resulting in two models: FLEL-ML-KNN and LE-ML-KNN. The performance of these models was systematically evaluated on each dataset using five commonly adopted multi-label metrics: AP、 HL、CV、RL and OE. The corresponding visualization results are presented in Fig. 3, with detailed quantitative results is provided *in Part E of the supplementary material.*

Experimental results demonstrate that FLEL-ML-KNN consistently outperforms LE-ML-KNN across most datasets and evaluation metrics. Notably, FLEL-ML-KNN achieves higher AP while significantly reducing HL, CV, RL, and OE.

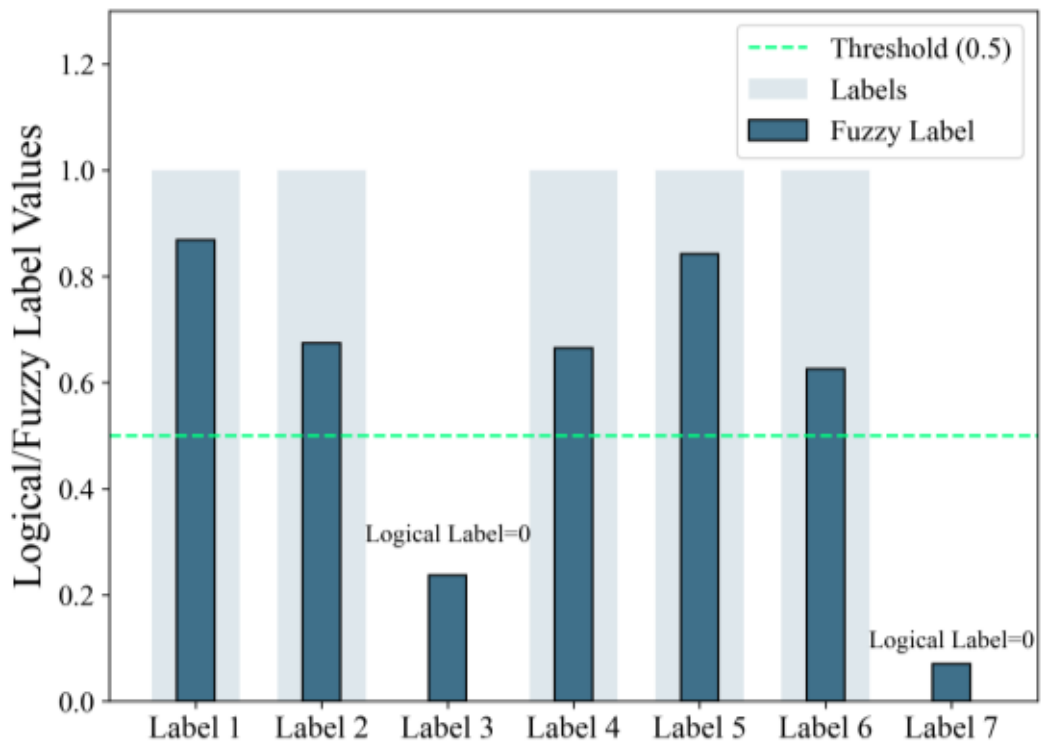

Fig. 4. Visualization of the logical labels under the dataset “flags” with the generated fuzzy labeled histograms.

This indicates that, compared to the traditional label enhancement method LE, which relies on the completeness assumption, the FLEL approach more accurately captures the intrinsic label ambiguity characteristic of real-world tasks in soft label modeling. Consequently, FLEL effectively uncovers latent information embedded within labels, thereby enhancing model generalization and robustness, making it more suitable for practical multi-label learning scenarios.

### e) *Fuzzy Labels Visualization*

To visually demonstrate the generation performance of the FL-Gen-LP method, we selected an instance from the *flags* dataset to compare the generated fuzzy labels with the original logical labels. The results are shown in Fig. 4. As illustrated in the figure, FL-Gen-LP can effectively generate fuzzy labels and accurately characterize the membership relationships between the instance and its labels. This not only validates the effectiveness of the FL-Gen-LP method in representing label uncertainty but also highlights its significant advantage in extracting fine-grained information.

## VII. CONCLUSION

To address the limitations of traditional label learning methods in handling data uncertainty and label ambiguity, this study introduces the concept of fuzzy labels for the first time and develops corresponding fuzzy label generation methods to better capture such uncertainty. Based on this foundation, two enhanced classification algorithms FLEL-SL-KNN and FLEL-ML-KNN are proposed, with KNN serving as the base model. Systematic experiments conducted on multiple datasets demonstrate that the proposed methods effectively improve the model's capacity to represent and learn from label information by incorporating fuzzy labels. These methods enable a more precise characterization of the complex relationships between labels and instances, thereby enhancing overall label learning performance.

Despite the significant advancements made in label learning achieved in this study, several areas warrant further exploration. For example, during the fuzzy label generation process, the selection of parameters can substantially impact the generated fuzzy labels, and in turn, affect overall learning performance. Future research could focus on developing more adaptive fuzzy label generation strategies to reduce reliance on manually tuned hyperparameters. Additionally, the computational complexity of the proposed FLEL-ML-KNN method remains a limitation. This is particularly pronounced in large-scale datasets. Further efforts are needed to develop optimized variants that retain performance while reducing computational overhead.

REFERENCE

[1] D. Li, L. L. Gong, S. L. Liu *et al.*, "Continual learning classification method with single-label memory cells based on the intelligent mechanism of the biological immune system," *J Intell Fuzzy Syst,* vol. 42, no. 4, pp. 3975-3991. 2022.

[2] H. I. Okagbue, O. A. Ijezie, P. O. Ugwoke *et al.*, "Single-label machine learning classification revealed some hidden but inter-related causes of five psychotic disorder diseases," *Heliyon,* vol. 9, no. 9, Sep. 2023.

[3] G. Tsoumakas, and I. Katakis, "Multi-Label Classification," *Int. J. Data Warehous. Min.*, pp. 1-13, 2007-07. 2007.

[4] M.-L. Zhang, and Z.-H. Zhou, "A Review on Multi-Label Learning Algorithms," *IEEE Trans. Knowl. Data Eng.,* vol. 26, pp. 1819-1837. 2014.

[5] M. Chandoo, "Logical labeling schemes," *Discrete Math,* vol. 346, no. 10, Oct. 2023.

[6] J. Liang, and D. Doermann, "Logical Labeling of document images using layout graph matching with adaptive learning," *Lect Notes Comput Sc,* vol. 2423, pp. 224-235. 2002.

[7] Y. Rangoni, A. Belaïd, and S. Vajda, "Labelling logical structures of document images using a dynamic perceptive neural network," *Int J Doc Anal Recog,* vol. 15, no. 1, pp. 45-55, Mar. 2012.

[8] Q. D. Lou, Z. H. Deng, Q. B. Sang *et al.*, "A Robust Multilabel Method Integrating Rule-Based Transparent Model, Soft Label Correlation Learning and Label Noise Resistance," *Ieee T Em Top Comp I,* vol. 8, no. 1, pp. 454-473, Feb. 2024.

[9] K. Lucke, A. Vakanski, and M. Xian, "Soft-Label Supervised Meta-Model with Adversarial Samples for Uncertainty Quantification," *Computers,* vol. 14, no. 1, Jan. 2025.

[10] J. Huang, C.-M. Vong, W. Qian *et al.*, "Online Label Distribution Learning Using Random Vector Functional-Link Network," *IEEE Trans. Emerg. Top. Comput. Intell.,* vol. 7, no. 4, pp. 1177-1190. 2023.

[11] T. Denoeux, and L. M. Zouhal, "Handling possibilistic labels in pattern classification using evidential reasoning," *Fuzzy Set Syst,* vol. 122, no. 3, pp. 409-424, Sep 16. 2001.

[12] X. Geng, "Label Distribution Learning," *IEEE Trans. Knowl. Data Eng.,* vol. 28, no. 7, pp. 1734-1748, Jul 1. 2016.

[13] F. Lin, and H. Ying, "Modeling and Control of Probabilistic Fuzzy Discrete Event Systems," *IEEE Trans. Emerg. Top. Comput. Intell.,* vol. 6, no. 2, pp. 399-408. 2022.

[14] N. N. Morsi, and M. M. Yakout, "Axiomatics for fuzzy rough sets," *Fuzzy Set Syst,* vol. 100, no. 1-3, pp. 327-342, Nov 16. 1998.

[15] T. M. Cover, and P. E. Hart, "Nearest neighbor pattern classification," *IEEE Trans. Inf. Theory,* vol. 13, pp. 21-27. 1967.

[16] M.-L. Zhang, and Z.-H. Zhou, "ML-KNN: A lazy learning approach to multi-label learning," *Pattern Recognit.*, pp. 2038-2048, 2007-07. 2007.

[17] O. Luaces, J. Díez, J. Barranquero *et al.*, "Binary relevance efficacy for multilabel classification," *Prog. Artif. Intell.*, pp. 303-313, 2012-12. 2012.

[18] C. Cortes, and V. Vapnik, "Support-Vector Networks," *Mach Learn,* vol. 20, no. 3, pp. 273-297, Sep. 1995.

[19] J. Read, B. Pfahringer, G. Holmes, and E. Frank eds., "Classifier Chains for Multi-label Classification," *Machine Learning and Knowledge Discovery in Databases,Lecture Notes in Computer Science*, 2009.

[20] K. Su, and X. Geng, "Soft Facial Landmark Detection by Label Distribution Learning," *Proc. AAAI Conf. Artif. Intell.*, pp. 5008-5015, 2019-09. 2019.

[21] X. Geng, K. Smith-Miles, and Z.-H. Zhou, "Facial age estimation by learning from label distributions," *Proc. AAAI Conf. Artif. Intell.*, pp. 451-456, 2022-09. 2022.

[22] D. ZHOU, X. Zhang, Y. Zhou *et al.*, "Emotion Distribution Learning from Texts," in: Proceedings of the 2016 Conference on Empirical Methods in Natural Language Processing, 2016.

[23] Y.-K. Li, M.-L. Zhang, and X. Geng, "Leveraging Implicit Relative Labeling-Importance Information for Effective Multi-label Learning," in: 2015 IEEE International Conference on Data Mining (ICDM), 2015.

[24] X. Wang, J. D. Peter, A. Slowik *et al.*, "Learning Fuzzy Label-Distribution-Specific Features for Data Processing," *IEEE Trans. Fuzzy Syst.,* vol. 33, no. 1, pp. 365-376, Jan. 2025.

[25] Z. Y. Wang, Z. X. Chen, C. Y. Liu *et al.*, "Diversity augmentation and multi-fuzzy label for semi-supervised semantic segmentation," *Neurocomputing,* vol. 630, May 14. 2025.

[26] T. L. Yang, C. Z. Wang, Y. Y. Chen, and T. Q. Deng, "A robust multi-label feature selection based on label significance and fuzzy entropy," *Int J Approx Reason,* vol. 176, Jan. 2025.

[27] Y.-K. Li, M.-L. Zhang, and X. Geng, "Leveraging Implicit Relative Labeling-Importance Information for Effective Multi-label Learning," *IEEE Int. Conf. Data Mining,* vol. vol. x, no. no. x, pp. 251-260. 2015.

[28] X. Zhu, and A. B. Goldberg, *Introduction to Semi-Supervised Learning*, 1st Edition ed., Cham, Switzerland: Springer Cham, 2009, pp. XII, 116.

[29] Y. Zhang, L. Liu, Q. Qiao, and F. Z. Li, "A Lie group Laplacian Support Vector Machine for semi-supervised learning," *Neurocomputing,* vol. 630, May 14. 2025.

[30] H. Y. Zhu, and X. Z. Wang, "A cost-sensitive semi-supervised learning modelbased on uncertainty," *Neurocomputing,* vol. 251, pp. 106-114, Aug 16. 2017.

[31] X. Zhu, and A. B. Goldberg, *Introduction to Semi-Supervised Learning*, 2009.

[32] G. Hinselmann, L. Rosenbaum, A. Jahn *et al.*, "Large-Scale Learning of Structure-Activity Relationships Using a Linear Support Vector Machine and Problem-Specific Metrics," *J Chem Inf Model,* vol. 51, no. 2, pp. 203-213, Feb. 2011.

[33] B. Shams, K. Reisch, P. Vajkoczy *et al.*, "Improved prediction of glioma-related aphasia by diffusion MRI metrics, machine learning, and automated fiber bundle segmentation," *Hum Brain Mapp,* vol. 44, no. 12, pp. 4480-4497, Aug 15. 2023.

[34] Z. W. Bian, Q. Chang, J. Wang *et al.*, "Takagi-Sugeno-Kang Fuzzy Systems for High-Dimensional Multilabel Classification," *IEEE Trans. Fuzzy Syst.,* vol. 32, no. 6, pp. 3790-3804, Jun. 2024.

# Supplementary Materials for the manuscript

# “Fuzzy Marking and Learning”

## Part A: The Concept of Fuzzy Labeling

The fuzzy labeling technique proposed in this paper is designed to overcome the limitations of traditional logical labels in expressing uncertainty. Compared to label distribution techniques, fuzzy labeling demonstrates greater adaptability to different scenarios. Their similarities and differences from the perspective of single-label and multi-label classifications are discussed as follows:

(1) **Single-Label Classification:** This category includes traditional binary classification and multi-class classification problems, where labels are completely mutually exclusive and each instance belongs to only one category. For example, in handwritten digit recognition, an instance can be assigned to only one digit class. In such settings, label distribution methods represent the degree of match between an instance and each possible label in a form similar to a probability distribution, ultimately assigning the instance to the category with the highest score. On the other hand, fuzzy labels represent the degree of attribution of an instance to each label using membership values. Similarly, the instance is assigned to the category with the highest membership value. Therefore, in single-label problems, both methods can effectively express the uncertain relationships between instances and labels while accurately enabling the final classification.

(2) **Multi-Label Classification:** This category pertains to traditional multi-label classification tasks. Within the label vector, there are no explicit relationships between different labels, and each instance can simultaneously belong to multiple labels. For example, a news article may strongly align with both the independent features of "politics" and "economy." In such cases, due to the completeness constraint, label distribution methods often causes interdependence among labels, which makes it hard for two independent labels to both show strong descriptive ability for the same instance. This limitation hinders the ability to represent each label independently. In contrast, fuzzy labels use membership degrees to independently describe the association of an instance with each label, free from the limitations of the completeness constraint. As a result, fuzzy labels let each label describe the instance independently, allowing label uncertainty to be fully expressed.

## Part B: Algorithmic Process

This paper introduces the concept of fuzzy labels, leveraging fuzzy set theory to define labels, thereby providing a powerful tool for representing label uncertainty. Furthermore, an efficient fuzzy label generation method, FL_Gen_LP, is proposed to effectively obtain fuzzy labels. This method derives fuzzy labels from the original data, capturing richer latent label information.

Building on this work, we take the classical single-label K-nearest neighbor algorithm (KNN) and multi-label KNN as examples to propose fuzzy label-enhanced single-label learning (FLEL_SL_KNN) and fuzzy label-enhanced multi-label learning (FLEL_ML_KNN) algorithms, respectively. The processes of these three algorithms are described below.

The algorithmic process of the fuzzy label generation algorithm FL_Gen_LP is described as Algorithm I:

**Algorithm I: FL_Gen_LP**

**Input:** dataset $\mathbf{S}=\left\{(\mathbf{x}_n,\mathbf{y}_n)\right\}_{n=1}^{N}$, $\mathbf{x}_n\in\mathcal{X}$ represents its *nth* instance, and $\mathbf{y}_n\in\mathcal{Y}$ is the logical label vector corresponding to the instance. $N$ denotes the number of instances, $K$ is the number of clusters, and $t$ represents the number of iterations.

**Output:** Fuzzy label $\mathbf{U}$

Procedure FL_Gen_LP:

1. Define a fully connected graph $\mathcal{G}=(V,E,\mathbf{W})$ to describe the relationships between instances in the dataset. Design a similarity measure function, denoted as: $w_{ij}^{New}=f_{close}\left(\mathbf{x}_i,\mathbf{x}_j\right)\times m_j^{k_{i,\max}}$.
   1). The spatial similarity between two instances is first measured using a Gaussian function.
   2). All instances are divided into $K$ clusters, resulting in a membership matrix $\mathbf{M}=\left[m_n^k\right]_{N\times K}$.
   3). For each instance $\mathbf{x}_i$, the cluster index $k_{\max}$ corresponding to its highest membership degree is selected.
   4). The similarity is then weighted based on the membership degree.
2. After obtaining the weight matrix $\mathbf{W}^{New}$, the classical iterative label propagation technique is used to learn the fuzzy labels $\mathbf{U}$ for the instances.

   1). Use the weight matrix $\mathbf{W}^{New}$ to derive the label propagation matrix $\mathbf{P}=\hat{\mathbf{A}}^{-\frac{1}{2}}\mathbf{W}\hat{\mathbf{A}}^{-\frac{1}{2}}\in\mathbb{R}^{N\times L}$.
   2). Initialize $\mathbf{U}^{(0)}=[y_{nl}]_{N\times L}$
   3). While not converged do
   4). $\mathbf{U}^{(t)}\leftarrow\alpha\mathbf{P}\mathbf{U}^{(t-1)}+(1-\alpha)\mathbf{Y}$;
   5). $t\leftarrow t+1$;
   6). Check the convergence conditions;
   7). End

Based on the above analysis, the prediction process of the FLEL-SL-KNN classifier is provided in Algorithm II:

**Algorithm II: FLEL-SL-KNN**

**Input:** A single-label dataset $\mathbf{S}=\{(\mathbf{x}_n,\mathbf{y}_n)\}_{n=1}^{N}$ and fuzzy labels $\mathbf{U}$, where $\mathbf{x}_n \in \mathcal{X} \subseteq \mathcal{R}^D$ represents the $nth$ instance, $\mathbf{y}_n \in \mathcal{Y} \subseteq \{0,1\}^C$ is the logical label vector corresponding to this instance, $D$ is the feature dimensionality, and $C$ is the number of classes.

**Output:** Fuzzy label **U**

Procedure FLEL-SL-KNN:

3. Calculate the distance $\mathcal{D}(\mathbf{x}_{test},\mathbf{x}_{train})$ between the test sample $\mathbf{x}_{test}$ and each training sample $\mathbf{x}_{train}$.
4. For the test sample $\mathbf{x}_{test}$, identify $K$ nearest neighbors and compute the inverse of the distances between the nearest neighbors and the test sample as weights $w_{\mathbf{x}_{Ne}\in\mathcal{N}_k(\mathbf{x}_{test})}$.
5. Perform a weighted average of the fuzzy labels of the nearest neighbors to obtain the final fuzzy label $u$ for the test sample.

Based on the above analysis, the training and prediction process of the FLEL-ML-KNN is provided in Algorithm III.

**Algorithm III: FLEL-ML-KNN**

**Input:** Train on the fuzzy-labeled dataset $S=\{\mathbf{x}_n,\mathbf{u}_n\}_{n=1}^{N_{train}}$, predict on the fuzzy-labeled dataset $S=\{\mathbf{x}_n,\mathbf{u}_n\}_{n=1}^{N_{test}}$, with the number of nearest neighbors $K$.

**Output:** Fuzzy label **U**

**1、Train Process:**

1) Compute the distance matrix $\mathcal{D}_{train}(\mathbf{x}_i,\mathbf{x}_j)$ between training samples.
2) Calculate the prior probability $P_{prior}(l)$ for each class.
3) Compute the conditional probabilities $P_{cond}$ (belonging to the class) and $P_{condN}$ (not belonging to the class).

**2、Prediction Process:**

1) Compute the distance $\mathcal{D}_{test}(\mathbf{x}_i,\mathbf{x}_j)$ between the test sample and the training samples.
2) Identify the nearest neighbors of the test sample in the training set and calculate the weighted sum $\mathcal{C}_{local}(l)$ of fuzzy labels for each class $l$ based on the labels of the nearest neighbors.
3) Use the theorem of Bayes to obtain the fuzzy label $P(l\,|\,\mathbf{x}_n)$ for class $l$.

# Part C: Synthetic Fuzzy Dataset Construction Method

The artificial dataset was constructed to visualize the effectiveness of the fuzzy markup generation method. A detailed description of the construction process for fuzzy single and multiple markers is presented.

Each data instance $\boldsymbol{x}=[x_1,x_2\cdots,x_5]\in\mathcal{R}^{1\times5}$ is defined as a five-dimensional feature vector to reduce dataset complexity. The corresponding fuzzy label $\boldsymbol{u}=[u_1,\ u_2,\ u_3]\in\mathcal{R}^{1\times3}$ is represented as a three-dimensional vector of fuzzy membership degrees.

## （1） *fuzzy single-labeled dataset construction*

The dataset consists of 1500 samples, each belonging to a single class, generated based on a three-cluster structure without overlap. Each class $k$ is represented by a randomly initialized cluster center $\mathbf{c}_k=\left[c_k^1,c_k^2,\cdots,c_k^5\right]\in\left[0,1\right]^5$, $k=1,2,3$. The feature vectors of samples in class $k$ are generated relative to this center. Specifically, each sample $\mathbf{x}$ is constructed as follows:

$$\mathbf{x}_i^k=\mathcal{N}\left(\mathbf{c}_k,\mathbf{\Sigma}_k\right),\ i\in\left\{1,2,\cdots,N_k\right\} \tag{1}$$

$$\mathbf{\Sigma}_k=\phi^2\left(\mathbf{I}+\rho\mathbf{A}_k\right) \tag{2}$$

where $\mathcal{N}\left(\mathbf{c}_k,\mathbf{\Sigma}_k\right)$ is a multivariate Gaussian distribution with mean $\mathbf{c}_k$ and covariance matrix $\mathbf{\Sigma}_k$. $N_k$ denotes the number of samples, satisfying $\sum_{k=1}^{3}N_k=1500$. $\mathbf{I}$ is the identity matrix, and $\mathbf{A}_k$ is a randomly generated symmetric positive definite matrix, representing the introduced feature correlations. $\phi$ is the scaling factor, used to control the overall magnitude of the covariance, and $\rho\in\left[0,1\right]$ is a weight parameter, used to adjust the influence of random correlations in the covariance matrix.

For a sample $\mathbf{x}_i^k$ belonging to the $k$-th cluster, its membership $u_k\left(\mathbf{x}_i\right)$ is calculated as the reciprocal of the Euclidean distance between the sample and the cluster center $\mathbf{c}_k$. The smaller the distance, the stronger the membership. The formula is as follows:

$$u_j\left(\mathbf{x}_i^k\right)=\begin{cases}\dfrac{1}{\left\|\mathbf{x}_i^k-\mathbf{c}_k\right\|_2^2}, & \text{if } j=k\\ 0, & \forall\ j\neq k\end{cases} \tag{3}$$

$$\mathbf{u}\left(\mathbf{x}_i^k\right)=\left[u_1\left(\mathbf{x}_i^k\right),u_2\left(\mathbf{x}_i^k\right),u_3\left(\mathbf{x}_i^k\right)\right] \tag{4}$$

Where $\mathbf{u}\left(\mathbf{x}_i\right)$ represents the final fuzzy single-label vector obtained for the sample $\mathbf{x}_i^k$, which reflects its membership across different categories.

## （2） *fuzzy multi-labeled dataset construction*

A fuzzy multi-label dataset with 1500 samples is constructed, where the samples are divided into three clusters with a certain degree of overlap and intersection. Each cluster has a designated center $\mathbf{c}_k=\left[c_k^1,c_k^2,\cdots,c_k^5\right]\in\left[0,1\right]^5$, and each data point is assigned fuzzy multi-labels based on its relative position to all clusters. The membership value for each category falls within the range $[0,1]$, and the total sum of membership values is $\sum_{k=1}^{3}u_k\neq1$. The detailed construction process is as follows:

The feature vectors of samples belonging to the $k$-th cluster are generated using Equation (1). To increase overlap and interaction between clusters, additional Gaussian noise is introduced to all samples:

$$\tilde{\mathbf{x}}_i^k = \mathbf{x}_i^k + \varepsilon, \quad \varepsilon = \mathcal{N}\left(0,\ \sigma^2\right) \tag{5}$$

where the variance of the Gaussian noise is set to $\sigma = 0.5$.

The fuzzy multi-label representation of sample $\tilde{\mathbf{x}}_i^k$ is obtained by computing its distance to each cluster center and transforming it using a Gaussian kernel function:

$$u_j\left(\tilde{\mathbf{x}}_i^k\right) = \exp\left(-\frac{\left\|\tilde{\mathbf{x}}_i^k - \mathbf{c}_k\right\|_2^2}{2\sigma^2}\right) + \alpha \tag{6}$$

$$\mathbf{u}\left(\tilde{\mathbf{x}}_i^k\right) = \left[u_1\left(\tilde{\mathbf{x}}_i^k\right), u_2\left(\tilde{\mathbf{x}}_i^k\right), u_3\left(\tilde{\mathbf{x}}_i^k\right)\right] \tag{7}$$

where $\mathbf{u}\left(\tilde{\mathbf{x}}_i^k\right)$ represents the final fuzzy multi-label vector obtained for sample $\tilde{\mathbf{x}}_i^k$ after computation.

# Part D: Evaluation Indicators

## (1) *single-marker evaluation indicators*

The experiment employs three commonly used classification evaluation metrics: Accuracy, F1-score, and the Area Under the Receiver Operating Characteristic Curve (ROC-AUC). Accuracy measures the overall correctness of the model's predictions. The F1-score provides a balanced assessment of the model's reliability, coverage, and class balance. ROC-AUC evaluates the classifier's ability to distinguish between positive and negative classes across different decision thresholds. The specific formulas are presented in Table S1.

TABLE S1.
THE BINARY CLASSIFICATION EVALUATION METRICS

| Measure | Formula | Mean |
|---|---|---|
| **Accuracy** | $Acc = \frac{TP+TN}{T+F}$ | The proportion of correctly predicted samples out of all samples. |
| **F1** | $\frac{2}{F1} = \frac{1}{P} + \frac{1}{R}$ | The harmonic means of precision and recall. |
| **ROC** | $ROC = 1 - R = \frac{FN}{T}$ | The proportion of false negatives among all positive samples. |
| **AUC** | $AUC = 1 - P = \frac{FP}{TP+FP}$ | The proportion of negative samples among the predicted positives. |

## (2) *multi-marker evaluation indicators*

To facilitate the description of evaluation metrics, the following definitions are introduced: Let $\boldsymbol{X}_{test} = \{(\boldsymbol{x}_{ti}, \boldsymbol{y}_{ti})\}_{i=1}^{N}$ represent an instance in the test set, where $\boldsymbol{y}_{ti} \in \{0,1\}^{l}$ denotes the true label set of $\boldsymbol{x}_{ti}$, and $\hat{\boldsymbol{y}}_{ti}$ represents the predicted label set. Define $\boldsymbol{L}_{ti}$ as the subset of labels relevant to instance $\boldsymbol{x}_{ti}$ and $\bar{\boldsymbol{L}}_{ti}$ as the subset of irrelevant labels. Let $f(\boldsymbol{x}_{ti}, l)$ denote the confidence score of the $l$th label belonging to instance $\boldsymbol{x}_{ti}$, and let $R(\boldsymbol{x}_{ti}, l)$ be the ranking function associated with $f(\boldsymbol{x}_{ti}, l)$, such that if $f(\boldsymbol{x}_{ti}, l') > f(\boldsymbol{x}_{ti}, l)$, then $R(\boldsymbol{x}_{ti}, l') < R(\boldsymbol{x}_{ti}, l)$. Based on these notations, Table S2 summarizes the five-evaluation metrics used in this study, along with their definitions and formulas, where "↑" indicates that a higher value corresponds to better classification performance, and "↓" indicates that a lower value corresponds to poorer classification performance.

TABLE S2.
THE MULTI-LABEL CLASSIFICATION EVALUATION METRICS

| Measure | Formula | Mean |
|---|---|---|
| Hamming Loss (HL) ↑ | $HL = \frac{1}{N}\sum_{i=1}^{N}\frac{\lvert \hat{y}_{ti} \oplus \boldsymbol{y}_{ti} \rvert}{L}$ | This metric evaluates the proportion of incorrectly predicted labels. |
| Average Precision (AP) ↓ | $AP = \frac{1}{N}\sum_{i=1}^{N}\frac{1}{\lvert L_{ti} \rvert}\sum_{l \in L_{ti}}\frac{\lvert \{ l' \mid R(\boldsymbol{x}_{ti}, l') \le R(\boldsymbol{x}_{ti}, l), l' \in L_{ti} \} \rvert}{R(\boldsymbol{x}_{ti}, l)}$ | This metric measures the average proportion of relevant labels ranked lower than a specific label. |
| One Error (OE) ↓ | $OE = \frac{1}{N}\sum_{i=1}^{N}\left[ [\arg\max_{l' \in L_{ti}} f(\boldsymbol{x}_{ti}, l')] \notin L_{ti} \right]$ | This metric assesses the proportion of relevant labels that are not correctly predicted within the top ranks of the predicted label set. |
| Ranking Loss (RL) ↓ | $RL = \frac{1}{N}\sum_{i=1}^{N}\frac{\lvert \{(y_1, y_2) \mid f(\boldsymbol{x}_{ti}, l) \le f(\boldsymbol{x}_{ti}, l'), (l, l') \in L_{ti} \times \bar{L}_{ti}\} \rvert}{\lvert L_{ti} \rvert \lvert \bar{L}_{ti} \rvert}$ | This metric evaluates the average proportion of errors in the ranking between relevant and irrelevant labels in the label set. |
| Coverage (CV) ↓ | $CV = \frac{1}{N}\sum_{i=1}^{N}\max_{l \in L_{ti}} R(\boldsymbol{x}_{ti}, l) - 1$ | This metric assesses the average number of times the relevant labels for an instance are found. |

## Part E: Performance Comparison of Different Soft Label Generation Methods

To evaluate the effectiveness of the proposed fuzzy label generation method (FLEL) against the existing label enhancement method (LE) in terms of latent information mining and model learning capability, this section presents a comparative study conducted on real-world datasets. Specifically, soft labels were generated on nine publicly available multi-label datasets using both methods. These soft labels were then employed to enhance the traditional multi-label classification algorithm ML-KNN, resulting in two models: FLEL-ML-KNN and LE-ML-KNN. The performance of these models was systematically evaluated on each dataset using five commonly adopted multi-label metrics: AP、 HL、CV、RL and OE. The comparative quantitative results are presented in Table S3.

**TABLE S3.**
**MEAN (SD) OF ALL PERFORMANCE METRICS FOR THE PROPOSED FLEL-ML-KNN AND LE-ML-KNN**

| Dataset | Method | Metric | | | | |
|---|---|---|---|---|---|---|
| | | AP↑ | HL↓ | OE↓ | RL↓ | CV↓ |
| Flags | LE-ML-KNN | 0.7946 (0.0237) | 0.4781 (0.0214) | 0.2860 (0.0547) | 0.2315 (0.0362) | 0.5421 (0.0290) |
| | FLEL-ML-KNN | **0.8165 (0.0123)** | **0.2563 (0.0001)** | **0.2754 (0.0312)** | **0.2107 (0.0142)** | **0.5311 (0.0217)** |
| CAL500 | LE-ML-KNN | **0.4984 (0.0145)** | 0.1497 (0.0031) | **0.1156 (0.0243)** | **0.1813 (0.0057)** | 0.7449 (0.0084) |
| | FLEL-ML-KNN | 0.4964 (0.0117) | **0.1300 (0.0125)** | 0.2388 (0.0115) | 0.1887 (0.0123) | **0.7421 (0.0124)** |
| Emotions | LE-ML-KNN | 0.6880 (0.0338) | 0.2847 (0.0054) | 0.3912 (0.0563) | 0.2969 (0.0417) | 0.4243 (0.0353) |
| | FLEL-ML-KNN | **0.8018 (0.0256)** | **0.1885 (0.0214)** | **0.2683 (0.0536)** | **0.1612 (0.0373)** | **0.2860 (0.0334)** |
| Yeast | LE-ML-KNN | 0.7102 (0.0058) | 0.3026 (0.0027) | 0.2491 (0.0061) | 0.2068 (0.0052) | 0.4871 (0.0034) |
| | FLEL-ML-KNN | **0.7761 (0.0113)** | **0.1842 (0.0124)** | **0.2317 (0.0148)** | **0.1634 (0.0186)** | **0.4351 (0.0133)** |
| Medical | LE-ML-KNN | 0.7419 (0.0020) | 0.0190 (0.0020) | 0.2944 (0.0474) | 0.0782 (0.0117) | 0.1052 (0.0130) |
| | FLEL-ML-KNN | **0.7954 (0.0103)** | **0.0157 (0.0142)** | **0.2671 (0.0190)** | **0.0431 (0.0134)** | **0.0733 (0.0166)** |
| Birds | LE-ML-KNN | 0.2492 (0.0173) | 0.0530 (0.0038) | 0.7953 (0.0223) | 0.2815 (0.0273) | 0.1902 (0.0310) |
| | FLEL-ML-KNN | **0.3550 (0.0156)** | **0.0362 (0.0150)** | **0.6515 (0.0119)** | **0.1817 (0.0122)** | **0.1246 (0.0134)** |
| Scene | LE-ML-KNN | **0.8569 (0.0109)** | **0.0886 (0.0040)** | **0.2310 (0.0156)** | **0.0904 (0.0083)** | 0.0908 (0.0072) |
| | FLEL-ML-KNN | 0.8549 (0.0112) | 0.0918 (0.0003) | 0.2368 (0.0146) | 0.0910 (0.0015) | **0.0904 (0.0001)** |
| Mirflickr | LE-ML-KNN | **0.4488 (0.0115)** | 0.1548 (0.0014) | 0.5951 (0.0251) | **0.2377 (0.0060)** | 0.4675 (0.0109) |
| | FLEL-ML-KNN | 0.4201 (0.0000) | **0.1406 (0.0003)** | **0.5865 (0.0107)** | 0.3280 (0.0186) | **0.4511 (0.0002)** |
| Rcv1 (subset1) | LE-ML-KNN | 0.3368 (0.0055) | 0.0284 (0.0004) | 0.6415 (0.0036) | 0.1516 (0.0018) | 0.2929 (0.0035) |
| | FLEL-ML-KNN | **0.5026 (0.0125)** | **0.0275 (0.0002)** | **0.5098 (0.0000)** | **0.0901 (0.0211)** | **0.1892 (0.0015)** |

## Part F: Experimental Comparison Visualization on Real-World Datasets

### （1） *FLEL-SL-KNN*

As shown in Table III of the main text, FLEL-SL-KNN significantly outperforms the conventional KNN on five real-world datasets. The visualization results are presented in Fig. S1.

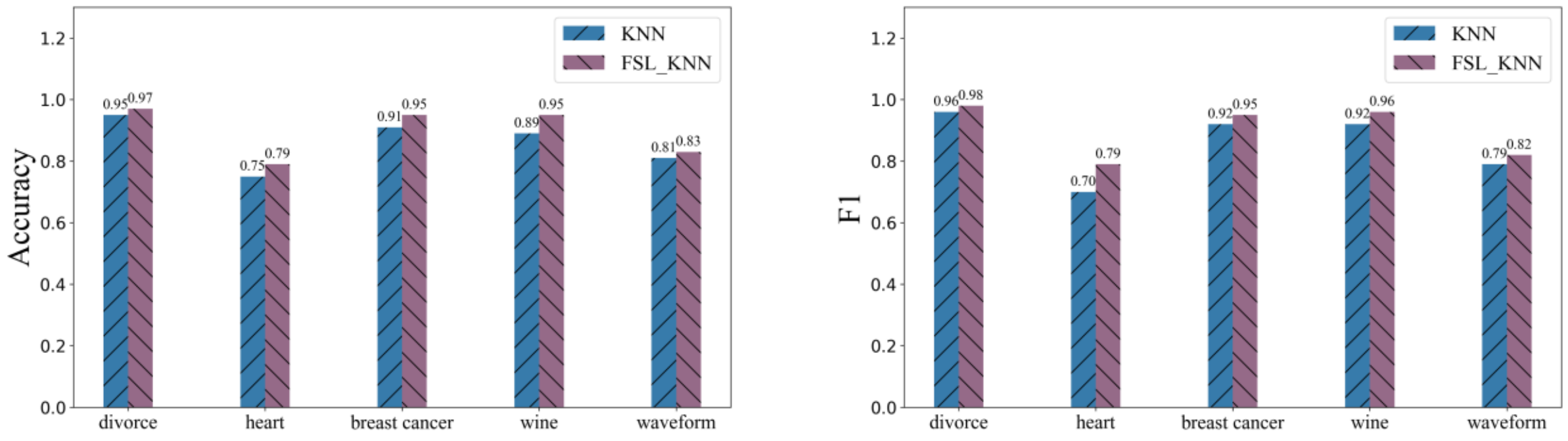

Fig. S1. Comparison of accuracy and F1 scores across real single-label datasets

### （2） *FLEL-ML-KNN*

Table VI in the main text presents the performance comparison between ML-KNN and FLEL-ML-KNN on nine real-world datasets. The visualization results are shown in Fig. S2.

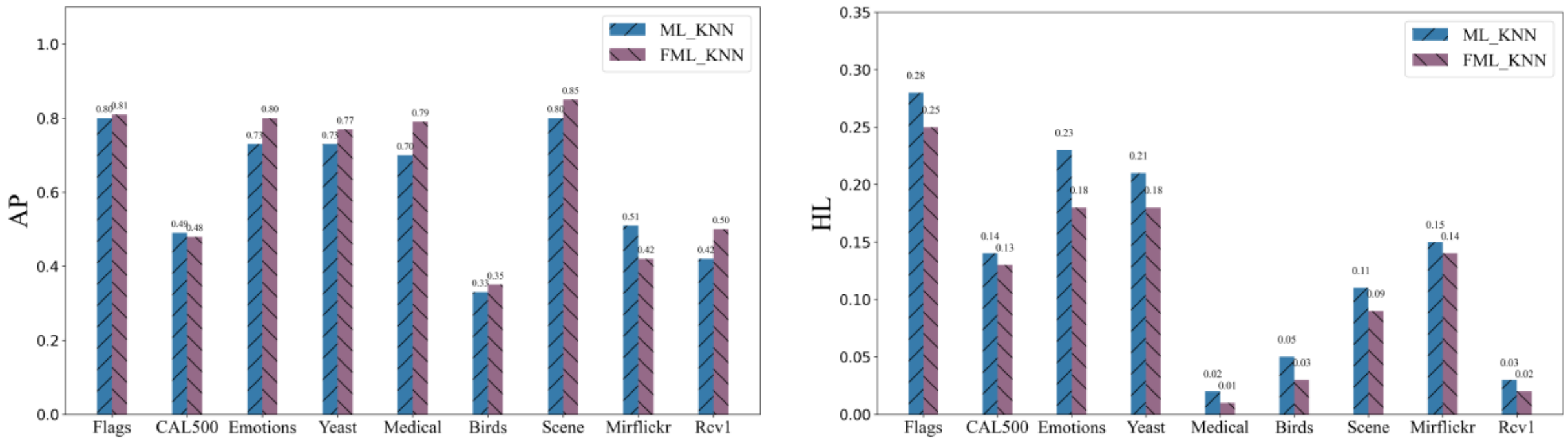

Fig. S2. Comparison of AP and HL Scores Across All Multi-Label Datasets

## Part G: Time Complexity Analysis

### （1） *FLEL-SL-KNN*

To comprehensively evaluate the time efficiency and practical applicability of the FLEL-SL-KNN method in single-label tasks, we compared the time complexity from three perspectives: the prediction time of KNN based on logical labels, the prediction time of FLEL-SL-KNN based on fuzzy labels, and the overall time encompassing fuzzy label generation and FLEL-SL-KNN prediction. The experimental results are shown in Fig. S3.

The results indicate that the prediction time based on fuzzy labels is slightly higher than that based on logical labels in single-label prediction tasks. Additionally, the fuzzy label generation process causes a small increase in total computation time, which is within an acceptable range and does not greatly affect

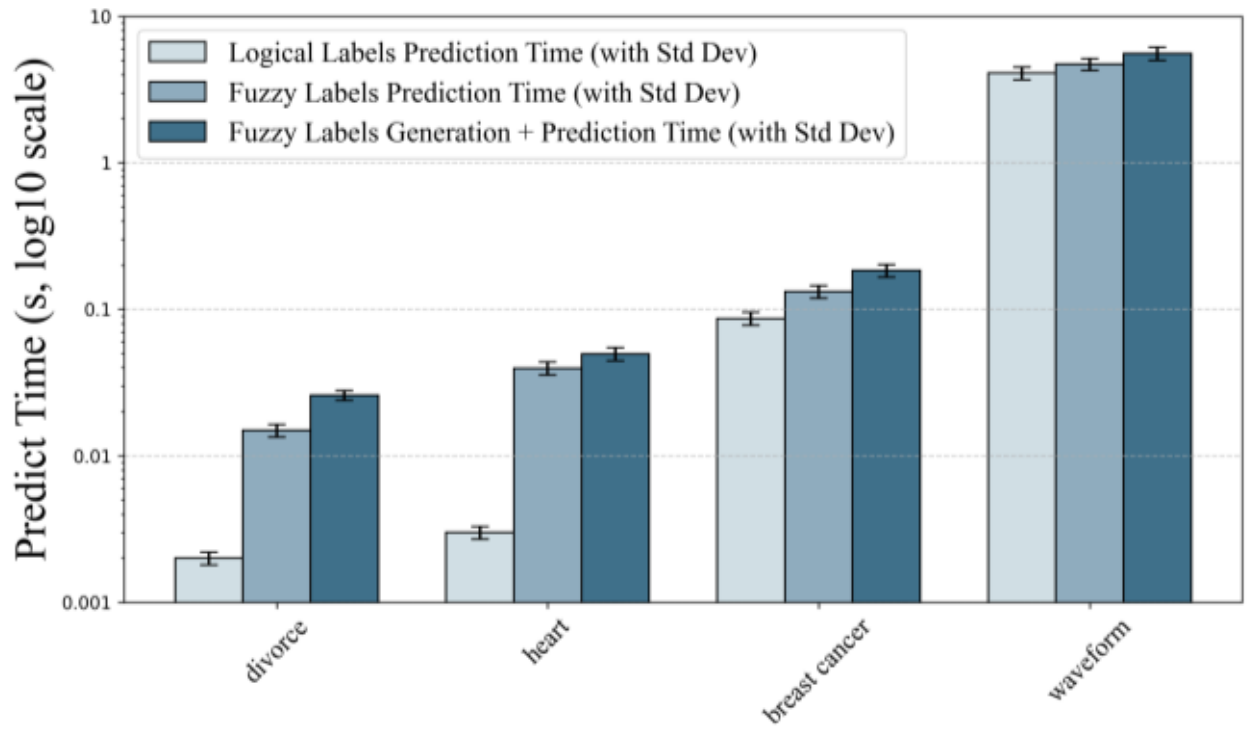

Fig. S3. Comparison of the differences between logical label prediction time, fuzzy label prediction time, and total fuzzy label generation and prediction time on each of the four single-labeled datasets

overall efficiency. In summary, the time cost of the FLEL-SL-KNN method in single-label tasks is similar to that of traditional logical label methods, which shows its high efficiency in practical use.

### （2） *FLEL-ML-KNN*

To comprehensively evaluate the time efficiency and practical performance of the FLEL-ML-KNN method in multi-label tasks, we compared its time complexity from three perspectives: the prediction time of ML-KNN based on logical labels, the prediction time of FLEL-ML-KNN based on fuzzy labels, and the overall time for fuzzy label generation and FLEL-ML-KNN prediction. The experimental results are shown in Fig. S4. The results indicate that in multi-label prediction tasks, the prediction time based on fuzzy labels is not significantly different from that based on logical labels, demonstrating that the computational cost of the FLEL-ML-KNN method is comparable to that of traditional logical label methods. Moreover, although the fuzzy label generation process slightly increases computation time, the increase is within an acceptable range and does not significantly impact overall computational efficiency. This demonstrates that FLEL-ML-KNN can effectively enhance the model's predictive capability while maintaining computational efficiency, ensuring its practicality.

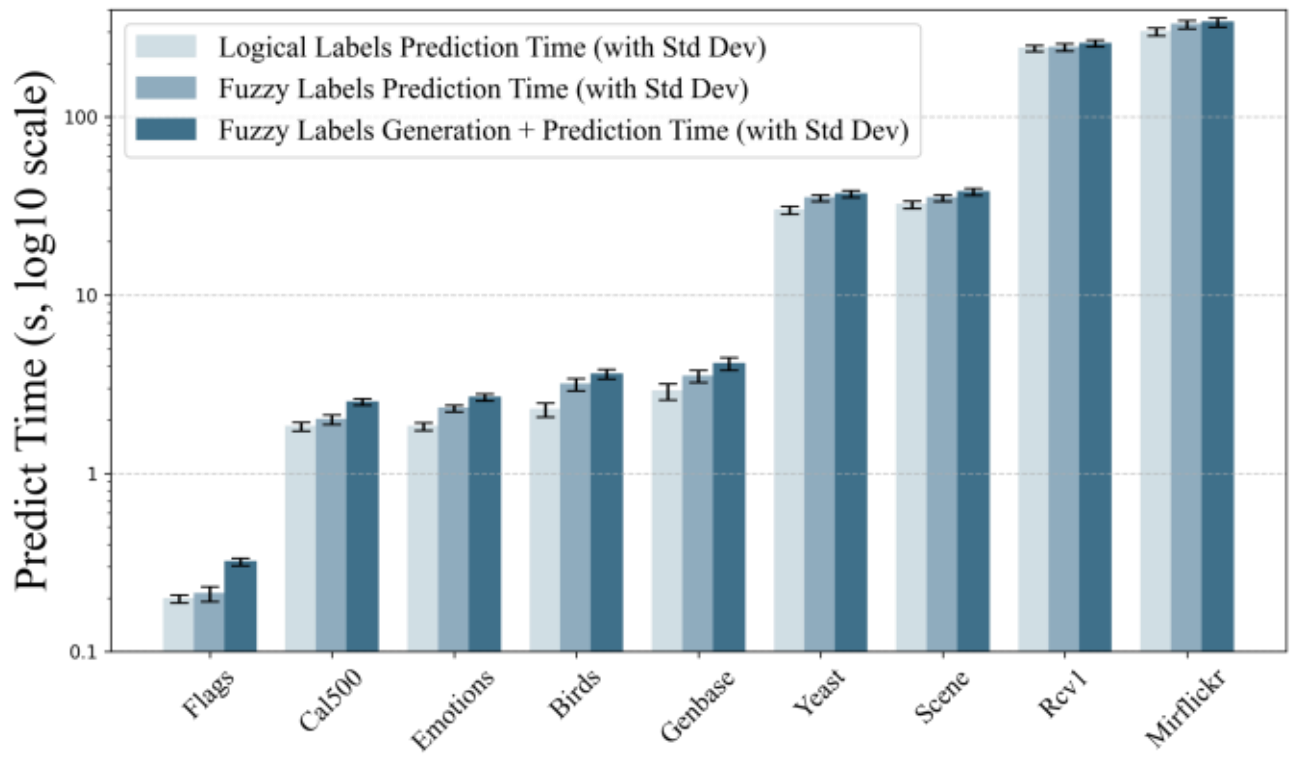

Fig. S4. Comparing the difference between logical label predictior